\documentclass[10pt,twocolumn,letterpaper]{article}
\usepackage[pagenumbers]{cvpr}
\usepackage[accsupp]{axessibility}
\usepackage[table]{xcolor}
\definecolor{rowgray}{gray}{0.92} 

\def\MethodName{POS-ISP}
\def\ActionNetName{sequence predictor} 
\def\ParamNetName{parameter predictor}

\definecolor{cvprblue}{rgb}{0.21,0.49,0.74}
\usepackage[pagebackref,breaklinks,colorlinks,allcolors=cvprblue]{hyperref}
\usepackage{booktabs}
\usepackage{multirow}
\usepackage{comment}

\title{
POS‑ISP: Pipeline Optimization at the Sequence Level for Task‑aware ISP
}

\author{Jiyun Won$^1$
\quad
Heemin Yang$^1$
\quad
Woohyeok Kim$^2$
\quad
Jungseul Ok$^{1,2}$
\quad
Sunghyun Cho$^{1,2}$ \vspace{-2mm}\\ \\
POSTECH CSE$^1$ \& GSAI$^2$\\
{\tt\small \{w1jyun, heeminid, woohyeok, jungseul, s.cho\}@postech.ac.kr}
}

\begin{document}
\maketitle
\begin{abstract}
Recent work has explored optimizing image signal processing (ISP) pipelines for various tasks by composing predefined modules and adapting them to task-specific objectives. However, jointly optimizing module sequences and parameters remains challenging. Existing approaches rely on neural architecture search (NAS) or step-wise reinforcement learning (RL), but NAS suffers from a training-inference mismatch, while step-wise RL leads to unstable training and high computational overhead due to stage-wise decision-making.
We propose \MethodName{}, a sequence-level RL framework that formulates modular ISP optimization as a global sequence prediction problem. Our method predicts the entire module sequence and its parameters in a single forward pass and optimizes the pipeline using a terminal task reward, eliminating the need for intermediate supervision and redundant executions. Experiments across multiple downstream tasks show that \MethodName{} improves task performance while reducing computational cost, highlighting sequence-level optimization as a stable and efficient paradigm for task-aware ISP. The project page is available at \href{https://w1jyun.github.io/POS-ISP}{https://w1jyun.github.io/POS-ISP}
\vspace{-0.4cm}
\end{abstract}    
\section{Introduction}
\label{sec:intro}
Image signal processors (ISPs) transform RAW sensor data captured by digital cameras into sRGB images suitable for human perception or machine vision. 
Conventional ISPs apply a fixed chain of operations such as white balance and tone mapping that are primarily designed to enhance image quality.
While such fixed pipelines are suitable for general photography, they often fail to align with the preferences or objectives of specific tasks, ranging from visual appearance optimization to high-level vision tasks such as object detection and semantic segmentation.
Although ISPs can be manually tuned by golden-eye experts, the process is time-consuming and difficult because it requires precise adjustment of many tightly coupled parameters for each task.
As a result, it is difficult to achieve consistent and optimal performance across different objectives.

To obtain enhanced ISP pipelines for downstream tasks, data-driven approaches have recently been proposed that learn ISPs directly from data. 
Among these, modular approaches have attracted particular attention due to their practical advantages. 
They decompose the ISP pipeline into well-established operations such as white balance and denoising, and optimize the pipeline in a task-driven manner. 
This modular design is especially appealing because the operations are already integrated into imaging systems and have low computational complexity, making them suitable for practical deployment~\cite{chen2018learning, chen2025task}.
However, despite this efficiency, optimizing modular ISPs remains difficult, since selecting the best sequence of modules and tuning parameters often requires non-differentiable search procedures.

To address this challenge, several approaches have recently explored neural architecture search (NAS) or reinforcement learning (RL) for modular ISP optimization~\cite{yu2021reconfigisp, shin2022drlisp, wang2024adaptiveisp}.
While they resolve the issue of non-differentiable optimization, they also introduce new limitations.
First, the NAS-based method~\cite{yu2021reconfigisp} enables gradient-based optimization by mixing the outputs of candidate modules.
However, the reliance on mixture training causes inconsistency at inference, where the modules are discretely selected.
Second, RL-based methods~\cite{shin2022drlisp, wang2024adaptiveisp} model ISP optimization as a stepwise RL formulation that performs sequential decision-making at each intermediate stage of the ISP pipeline.
Unfortunately, such a formulation requires repeated evaluations and relies on future reward estimation, resulting in unstable training and  high computational overhead.
This instability is a well-known issue in deep reinforcement learning, arising from the difficulty of stabilizing bootstrapped value estimation under function approximation~\cite{fujimoto2018addressing,van2018deep}.
Moreover, since the decision process must be repeatedly evaluated at each stage to determine the next action, this stepwise formulation is structurally inefficient.

In this paper, we present \emph{\MethodName{}}, a novel RL framework for searching optimal modular ISP pipelines tailored to downstream tasks. Unlike existing RL methods that make stepwise decisions, \MethodName{} performs sequence-level optimization by evaluating the entire pipeline with a single final reward. This formulation enables direct evaluation of the final result, avoiding unstable future reward estimation and leading to more stable optimization. It also captures dependencies between ISP modules, allowing the policy to consider the global pipeline context when predicting module sequences. Furthermore, \MethodName{} predicts the entire pipeline in a single forward pass, significantly reducing memory and computation. Such efficiency is essential for ISPs deployed on mobile or edge devices, where they must function as lightweight pre-processing components.

To enable sequence-level, context-aware optimization, \MethodName{} adopts a carefully designed network to predict the module sequence, named \emph{\ActionNetName{}}, along with \emph{\ParamNetName{}} for predicting module parameters.
The \ActionNetName{} is a recurrent policy network that predicts the entire module sequence by leveraging contextual information from preceding modules.
At each recurrent step, the \ActionNetName{} takes the previously selected module along with the hidden state, which contains contextual information of preceding modules, and predicts a probability distribution over the module candidates.
Thanks to this context-aware and lightweight recurrent design, \MethodName{} can predict the module sequence with reduced computational cost and memory overhead while considering the dependencies between the modules.
In parallel, \ParamNetName{} predicts module parameters with a small encoder–decoder network conditioned on the input image, enabling image-adaptive parameter prediction.
The predicted module sequence and its corresponding parameters together form a complete ISP pipeline, whose output image is evaluated based on task-driven performance.

We validate \MethodName{} by measuring its task-specific performance after optimization for multiple tasks, including object detection, instance segmentation, and image enhancement.
Extensive experiments demonstrate that \MethodName{} outperforms other task-aware ISP optimization methods both quantitatively and qualitatively, with a lower computational cost and memory footprint.

Our main contributions can be summarized as follows:
\begin{itemize}
   \item We introduce \MethodName{}, a framework that performs sequence-level optimization of the ISP pipeline by predicting the entire pipeline in a single forward pass, directly optimizing the final task reward without relying on unstable stepwise supervision.
   \item We design a recurrent \ActionNetName{} that enables sequence-level prediction while capturing inter-module dependencies for context-aware optimization.
   \item We evaluate \MethodName{} on object detection, instance segmentation, and image enhancement. Extensive experiments demonstrate that \MethodName{} achieves state-of-the-art performance with substantially reduced computational cost and memory usage.
\end{itemize}

\section{Related Work}
\label{sec:related}
With the advancement of deep learning, several works have been proposed to replace conventional ISPs with end-to-end deep neural networks~\cite{kim2024paramisp, afifi2021ciexyznet, brooks2019upi, conde2022model, xing2021invisp, zamir2020cycleisp}.
They aim to design neural networks that learn RAW-to-RGB mappings.
Thanks to strong image priors, they have shown promising performance in mimicking the ISPs.

Beyond directly learning RAW-to-RGB mappings, several works have explored optimizing ISP configurations for downstream tasks~\cite{tseng2019hyperparameter, qin2022attention}.
Tseng \etal~\cite{tseng2019hyperparameter} 
train a differentiable proxy to approximate a black-box ISP and then optimize the ISP hyperparameters through this proxy to maximize downstream task performance.
Motivated by this work, Qin \etal~\cite{qin2022attention} introduced an attention-aware framework to better capture the important image regions when predicting ISP parameters.
{However, these approaches rely on neural networks to approximate ISP behavior or predict its parameters, which increases computational complexity.}

In contrast, modular ISP designs have practical advantages in terms of interpretability and computational efficiency.
Therefore, there have been several works that optimize modular ISP pipelines in a task-driven manner~\cite{shin2022drlisp, yu2021reconfigisp, wang2024adaptiveisp}.
To resolve the non-differentiability of optimizing the ISP module sequence and parameters, ReconfigISP~\cite{yu2021reconfigisp} employs differentiable proxy networks to approximate ISP modules, enabling gradient-based optimization of the sequence and parameters.
During the architecture search, it assigns learnable weights to the modules at each step and mixes the module outputs based on the weights for differentiability.
It then selects the module with the highest weight to construct the ISP pipeline for inference.
{However, the mismatch between soft selection during search and hard selection during inference leads to suboptimal performance.}

Other approaches cast the ISP search problem as a sequential decision-making process and adopt a reinforcement learning (RL) framework.
DRL-ISP~\cite{shin2022drlisp} sequentially selects ISP modules to construct the pipeline, with a search space that includes both CNN-based modules and discretized variants of traditional ISP operators. AdaptiveISP~\cite{wang2024adaptiveisp} further extends this framework by searching module sequences in a discrete space while predicting module parameters in a continuous space.
However, despite these advances, prior RL-based ISP search methods rely on an actor–critic framework, where a critic network estimates future rewards to guide the agent's decisions. This reliance on intermediate supervision often leads to suboptimal performance due to unstable critic optimization and also incurs substantial computational overhead, as decisions are made sequentially at each stage.

Unlike these stepwise RL approaches, our method performs sequence-level optimization in a single forward pass without intermediate supervision. This results in more stable training and improved computational efficiency.
\section{\MethodName{}}

\subsection{Problem Formulation}

\begin{figure*}[t]
    \centering
    \includegraphics[width=0.93\linewidth]{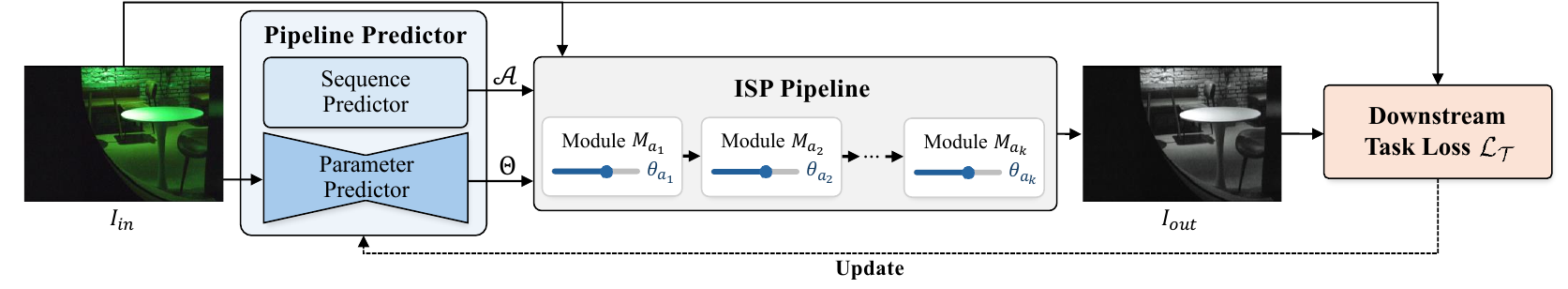}
    \vspace{-0.3cm}
    \caption{\textbf{Overview of the proposed method.} \MethodName{} aims at constructing the ISP pipeline that best performs for the downstream task. The \ActionNetName{} predicts the image processing module sequence based on the learned policy, and the \ParamNetName{} estimates the corresponding parameters of each module.
    }
    \vspace{-0.3cm}
    \label{fig:main}
\end{figure*}
\begin{figure}[t]
    \centering
    \vspace{0.2cm}
    \includegraphics[width=0.95\linewidth]{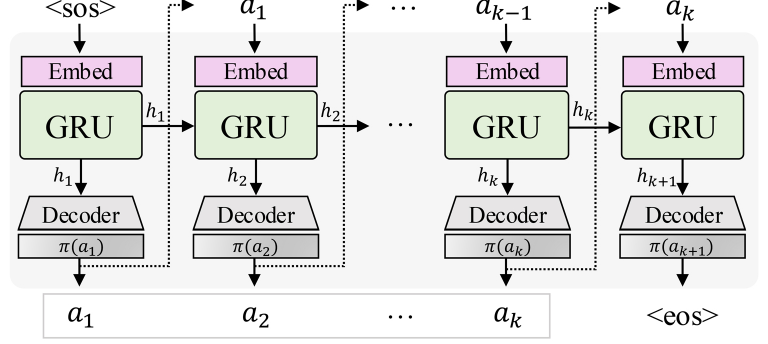}
    \caption{\textbf{Detailed architecture of sequence predictor.} The \ActionNetName{} predicts the image processing module sequence based on the learned policy.
    }
    \vspace{-0.4cm}
    \label{fig:estimator}
\end{figure}
\cref{fig:main} shows the overall framework of \MethodName{}.
The goal of \MethodName{} is to discover an ISP pipeline that transforms a RAW image $I_\text{in}$ into an sRGB image $I_\text{out}$, such that the output maximizes the performance of a target downstream task $\mathcal{T}$.
Following conventional camera ISPs, our framework models an ISP pipeline as a sequence of image processing modules, each with its own internal parameters.
In our experiments, we adopt the same set of modules as those used in AdaptiveISP~\cite{wang2024adaptiveisp}, encompassing standard ISP operations such as white balance and tone mapping. 
The detailed list of modules is provided in the supplementary material.

Formally, we define a candidate set of ISP modules $\mathbb{M}=\{\mathcal{M}_1, \cdots, \mathcal{M}_n\}$, where $n$ denotes the total number of available modules, and each module $\mathcal{M}_i$ is parameterized by its own parameters $\theta_i$.
Based on this, we model the ISP pipeline of \MethodName{} as:
\begingroup
\small
\begin{equation}
\begin{aligned}
I_{\text{out}}
&= \bigl(
\mathcal{M}_{a_k}(\cdot;\theta_{a_k})
\circ \cdots \circ
\mathcal{M}_{a_2}(\cdot;\theta_{a_2})
\circ
\mathcal{M}_{a_1}(\cdot;\theta_{a_1})
\bigr)\!(I_{\text{in}})
\\[-2pt]
&= F(I_{\text{in}} \,;\, \mathcal{A}, \Theta)
\end{aligned}
\label{eq:pipeline_comp}
\end{equation}
\endgroup
where $a_i$ corresponds to a module index such that $a_i \in \{1, ..., n\}$, and $k$ denotes the number of modules in the pipeline.
We denote the sequence of modules as $\mathcal{A} = (a_1, ..., a_k)$, and the corresponding module parameters are represented as $\Theta = (\theta_{a_1}, ..., \theta_{a_k})$.
For stability and tractability, we assume that each module is sampled at most once, i.e., $a_i \neq a_j$ if $i \neq j$.

With these definitions in place, our framework aims to find an optimal module sequence $\hat{\mathcal{A}}$ and the corresponding parameters $\hat{\Theta}$.
Specifically, we design our framework to determine the optimal sequence at the task level and the optimal parameters at the image level. 
In other words, our framework finds a single sequence $\hat{\mathcal{A}}$ for a given downstream task $\mathcal{T}$ and then uses the sequence across all images for that task, while training a parameter predictor network that predicts parameters adapted to each input image.

This separation is motivated by two key considerations.
First, conventional camera ISPs adopt a fixed sequence of modules (e.g., white balancing, tone mapping), with parameters that adapt to each image. 
This design reflects practical hardware constraints, as pipeline structures are typically embedded in silicon or firmware and cannot be reconfigured per image. By following this paradigm, our framework both aligns with real-world ISP design principles and enables the discovery of task-optimized pipelines that can be readily deployed in hardware.
Second, the order of operations in an ISP pipeline is largely determined by the target downstream task.
For example, tasks that rely on structural information (e.g., object detection) tend to place contrast and sharpening earlier in the sequence, whereas tasks targeting perceptual quality prioritize exposure and tone adjustments to achieve balanced brightness and color.

To find the optimal sequence $\hat{\mathcal{A}}$ and parameters $\hat{\Theta}$, our framework introduces two complementary components: \ActionNetName{} and \ParamNetName{}. 
The \ActionNetName{} models a probability distribution over possible ISP operation sequences, while the \ParamNetName{} predicts image‑adaptive parameters conditioned on an input image. 
Instead of directly committing to a single sequence, we model a distribution to capture the inherent uncertainty and variability in pipeline design, enabling exploration of multiple candidate structures during training and preventing premature convergence to suboptimal solutions.

By jointly training these networks for a downstream task $\mathcal{T}$, we simultaneously learn the distribution of task‑specific module sequences and image-adaptive parameters. 
After training, we select the most probable sequence for the target task and employ \ParamNetName{} to generate parameters tailored to each input image.

\subsection{Network Architecture}

\MethodName{} constructs its task-adaptive ISP pipeline by predicting a task-specific module sequence from the \ActionNetName{} and image-specific module parameters from the \ParamNetName{}.
The \ActionNetName{} models the full distribution over module sequences instead of predicting only the most probable pipeline. 
This probabilistic formulation enables thorough exploration of diverse yet plausible pipelines during ISP search.
The \ParamNetName{} predicts the image-specific module parameters given an input image.
In the following, we describe the networks in detail.
\vspace{-15pt}
\paragraph{Sequence predictor}
For an ISP sequence $\mathcal{A}=(a_1,\cdots,a_T)$, the \ActionNetName{} models its probability as:
\begingroup
\small{
\begin{equation}
    p(\mathcal{A}) = \prod_{i=1}^{T} p(a_i \mid a_{<i}),
\end{equation}
}
\endgroup
\par\noindent where $a_{<i} = (a_0, \cdots, a_{i-1})$, with $a_0=\texttt{<sos>}$ denoting a special token that represents the start of the sequence.
The sequence also includes a special token \texttt{<eos>} that terminates the sequence to allow pipelines of arbitrary length.
To parameterize this distribution, the \ActionNetName{} adopts a recurrent architecture based on Gated Recurrent Units (GRUs)~\cite{cho2014gru}, which are widely used in sequence modeling across domains such as natural language processing and recommendation systems, thanks to their efficiency and ability to capture sequential dependencies~\cite{serban2016building, hidasi2015session, ren2019repeatnet}.

\cref{fig:estimator} shows an overview of the \ActionNetName{}.
At the $i$-th recurrent step, the \ActionNetName{} takes the previous module index $a_{i-1}$ as input and embeds it into a vector.
Then, the GRU updates the hidden state $h_i$ using this embedding together with the previous hidden state $h_{i-1}$ (initialized as zeros). 
Here, $h_i$ encodes the past context $a_{<i}$ and serves as the sufficient representation for predicting $a_i$.
Given $h_i$, an MLP-based decoder followed by a softmax layer predicts the probability distribution $\pi(a_i)$ over candidate modules, parameterizing the conditional distribution $p(a_i \mid a_{<i})$ based on the hidden state $h_i$.

During the ISP search, we train the \ActionNetName{} by sampling ISP pipelines from the learned policy and evaluating their task performance.
At each recurrent step, $a_i$ is sampled from the probability distribution $\pi(a_i)$ and fed into the next step until the $\texttt{<eos>}$ token is produced, forming a complete pipeline.
To balance exploration and exploitation during policy training, we apply a temperature-controlled sampling strategy~\cite{cesa2017boltzmann}, encouraging exploration in the early phase and progressively focusing on exploitation in later stages.
After the search, the final ISP pipeline is generated using greedy decoding, where the highest-probability module is sequentially selected from $\pi(a_i)$ until $\texttt{<eos>}$ is reached.
Further details are provided in the supplementary material.

\vspace{-10pt}
\paragraph{Parameter predictor}
The \ParamNetName{} predicts the module parameters $\Theta$ conditioned on the input image.
A lightweight convolutional neural network (CNN)-based encoder processes a downsampled 64×64 input and extracts a compact feature representation, which is then passed through a decoder to produce the parameter sets for all modules in $\mathbb{M}$.
When constructing the ISP pipeline, only the parameters corresponding to the selected sequence $\mathcal{A} = (a_1, \dots, a_k)$ are applied; that is, ${\theta_{a_1}, \dots, \theta_{a_k}}$ are retrieved from $\Theta$ to form the final pipeline.

Although the parameter predictor can be conditioned on both the input image and the predicted module sequence, we empirically found that using only the image yields better performance. 
This is likely because sequence conditioning increases learning complexity, while image-only input acts as regularization. 
Importantly, as the sequence policy gradually converges toward high-performing pipelines, the parameter predictor, trained via task-driven feedback, naturally adapts to these dominant sequences. 
Even without direct access to the sequence, it learns to produce compatible parameters for frequently selected pipelines, enabling effective coordination and near-optimal performance.

\subsection{ISP Search}

Building on the \ActionNetName{} and parameter predictor, we search for effective ISP pipelines tailored to a downstream task $\mathcal{T}$. 
During ISP search, both predictors are jointly trained, enabling the selection of the most probable pipeline and its associated parameters after search. 

We formulate ISP search as a RL problem over discrete module sequences, while learning module parameters via differentiable optimization.
An ISP pipeline is represented by a module sequence $\mathcal{A}$ together with its parameter set $\Theta$, 
yielding $F(\cdot; \mathcal{A}, \Theta)$ that maps an input image $I_\text{in}$ to an output image $I_\text{out}$. 
Pipeline quality is assessed by applying $F$ to $I_\text{in}$ and measuring task performance on the output, 
which serves as the reward signal in the RL framework. 
Unlike a standard Markov decision process, our formulation does not involve explicit states or stepwise decisions: 
the \ActionNetName{} generates the module sequence in a single forward pass, 
while the \ParamNetName{} predicts the corresponding parameters conditioned on $I_\text{in}$. 
This design provides a terminal reward based on the performance of a fully-formed ISP, 
avoiding unstable reward estimation and enabling end-to-end optimization with improved stability.

We define the reward as the improvement in downstream task performance compared to the baseline input, with an additional penalty term to discourage degenerate solutions:
\begingroup
\small{
\begin{equation}
    \label{eq:reward}
    R(I_\text{in}, \mathcal{A}, \Theta) = \mathcal{L}_\mathcal{T}(I_\text{in}) - \mathcal{L}_\mathcal{T}(I_\text{out}) - P(I_\text{out}),
\end{equation}}
\endgroup
\par\noindent where $I_\text{in}$ is the input image, and $I_\text{out}=F(I_\text{in};\mathcal{A},\Theta)$ is the output processed by the ISP pipeline defined by $\mathcal{A}$ and $\Theta$. 
Here, $\mathcal{L}_\mathcal{T}$ denotes the loss function of the target task $\mathcal{T}$, and $P$ is a penalty term that prevents degenerate outputs. 
In the RL formulation, $I_\text{in}$ corresponds to the current state, while the choice of module sequence $\mathcal{A}$ and parameters $\Theta$ constitutes the action. 
The ISP pipeline $F$ represents the environment transition, producing the next state $I_\text{out}$. 
The reward $R$ thus measures how much the chosen action improves task performance relative to the baseline, while penalizing implausible results.

For example, when the task is object detection, $\mathcal{L}_\mathcal{T}$ is defined as the detection loss from a pretrained detector. 
In this case, the reward reflects the improvement in detection accuracy achieved by the ISP pipeline. 
More generally, $\mathcal{L}_\mathcal{T}$ can be adaptively defined for other tasks, allowing \MethodName{} to tailor its pipeline to diverse objectives by directly optimizing task-relevant metrics.

For the penalty term $P$, {in the detection and instance segmentation experiments}, we adopt the intensity-based penalty from the truncation condition in AdaptiveISP~\cite{wang2024adaptiveisp},
which discourages extreme pixel values:

\begingroup
\small{
\begin{equation}
P = \alpha_1 [I_\text{low} - \bar{I}_{\text{out}}]_+ + \alpha_2 [\bar{I}_{\text{out}} - I_\text{high}]_+ ,
\end{equation}}
\endgroup
\par\noindent where $\bar{I}_\text{out}$ is the mean intensity of $I_\text{out}$, and $I_\text{low}$ and $I_\text{high}$ are the lower and upper bounds, set to $0.01$ and $0.9$ following AdaptiveISP. Here, $[x]_+$ is equivalent to $\max(0, x)$. {This term serves as a soft regularizer that discourages extreme exposure shifts while preserving flexibility for optimization.}

Given the reward, we train the \ActionNetName{} and \ParamNetName{} alternately.
For the \ActionNetName{}, we update it via the REINFORCE policy gradient method~\cite{williams1992simple} to increase the likelihood of high-reward module sequences. 
Specifically, we define the learning objective for the \ActionNetName{} as
\begingroup
\small{
\begin{equation}
    \mathcal{L}_\text{seq} = -\hat{\mathbb{E}}_{\mathcal{A}\sim\pi}\left[R(I_\text{in},\mathcal{A}, \Theta)\,\cdot\,\sum_{i=1}^{k}\log\pi(a_i)\right],
\end{equation}
}
\endgroup
\par\noindent
where $\hat{\mathbb{E}}$ denotes the expectation over a mini-batch and $\pi$ denotes the probability of selecting $a_i$ at step $i$.
$\Theta$ is computed from $I_\text{in}$ using the \ParamNetName{}, i.e., $\Theta=\Theta(I_\text{in})$.
This objective encourages the \ActionNetName{} to assign higher probability to actions that yield higher rewards.
On the other hand, the parameter predictor is trained by minimizing the following loss via backpropagation:
\begingroup
\small{
\begin{equation}
\mathcal{L}_{\text{param}}
= \mathcal{L}_{\mathcal{T}}\!\left(I_\text{out}\right) + P(I_\text{out}),
\label{eq:param_loss_final}
\end{equation}}
\endgroup
which is equivalent to maximizing the reward in \cref{eq:reward}. 

After the search, we construct the final ISP pipeline as follows. 
First, the optimal module sequence $\hat{\mathcal{A}} = (\hat{a}_1, ..., \hat{a}_k)$ is obtained from the trained \ActionNetName{} by discretely selecting the module with the highest probability at each step until \texttt{<eos>} is reached. 
This sequence $\hat{\mathcal{A}}$ is fixed during inference. 
In parallel, the \ParamNetName{} takes the input image $I_\text{in}$ and predicts parameters for all $n$ modules $(\hat{\theta}_1, ..., \hat{\theta}_n)$. 
From these, the parameters corresponding to the selected sequence, $\hat{\Theta} = (\hat{\theta}_{a_1}, ..., \hat{\theta}_{a_k})$, are selected. 
Finally, the pipeline defined by $\hat{\mathcal{A}}$ and $\hat{\Theta}$ processes $I_\text{in}$ to produce the task-adapted output $I_\text{out}$.
\section{Experiments}
\label{sec:experiments}
\begin{figure*}[t]
    \centering
    \includegraphics[width=\linewidth]{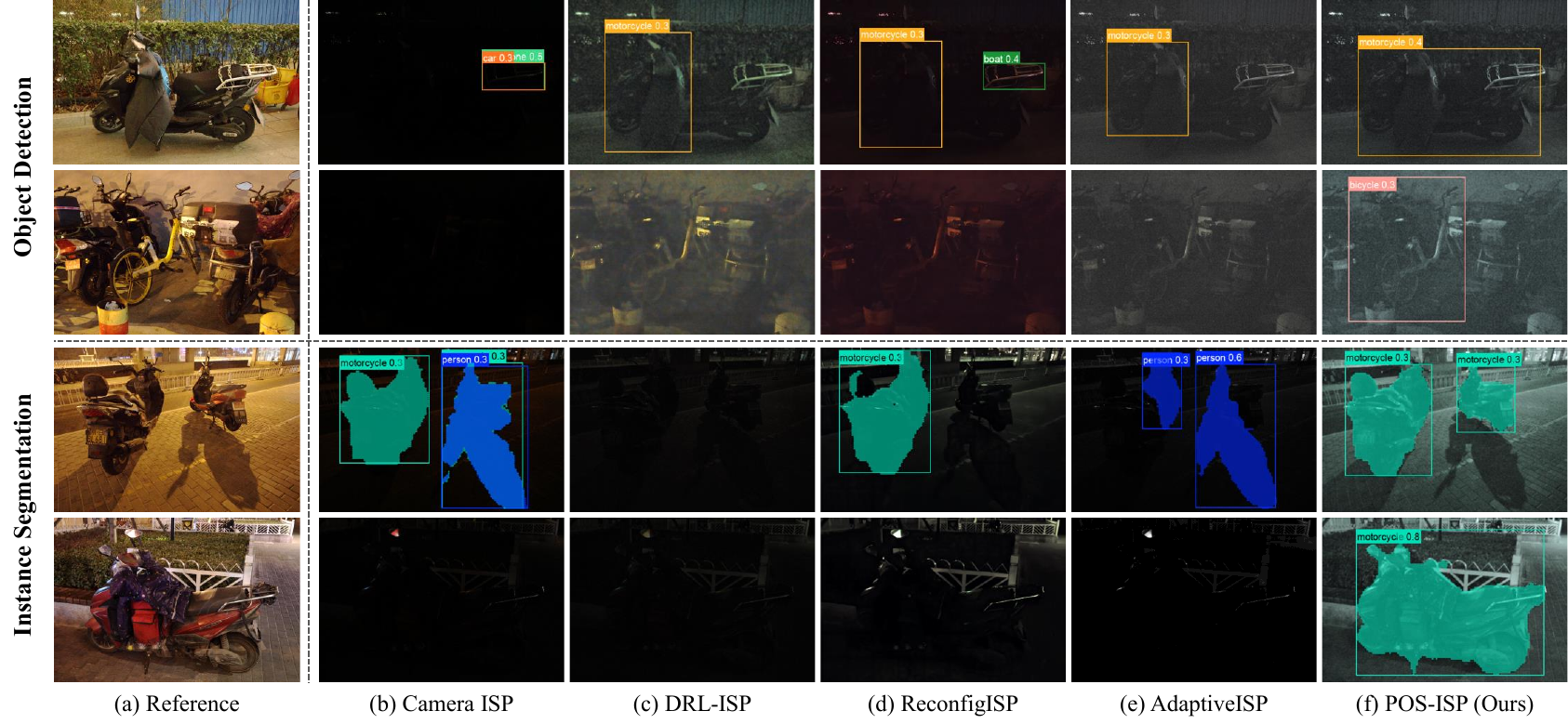}
    \vspace{-0.7cm}
    \caption{\textbf{Comparison of different ISP methods on object detection and instance segmentation tasks.} Reference images are well-lit scenes from the LOD and LIS datasets, with brightness increased by $1.5\times$ for visualization. More results are in the supplementary material.
    }
    \vspace{-0.1cm}
    \label{fig:comp_det_seg}
\end{figure*}
\paragraph{Implementation details}
We adopt the same candidate set of ISP modules as in AdaptiveISP~\cite{wang2024adaptiveisp}. 
Following their setting, we also assume that the input images are not completely raw sensor data, but have instead undergone only minimal preprocessing steps, such as defective pixel correction, black level correction, and demosaicking.
These operations are standard prerequisites in camera pipelines, applied prior to any higher-level ISP modules, ensuring that our framework operates on minimally processed inputs while remaining aligned with prior work.
Details on the ISP modules are provided in the supplementary material.

We implemented \MethodName{} using PyTorch.
We train the framework on resized image patches of size $512\times 512$ with a batch size of 8 for 15,000 iterations.
We use the Adam optimizer~\cite{kingma2014adam} with $\beta_1 = 0.9$ and $\beta_2 = 0.99$, and the learning rates are set to $1\times10^{-4}$ for the \ParamNetName{} and $3\times10^{-5}$ for the \ActionNetName{} with no learning rate scheduling.
The training process takes approximately 24 hours on a single RTX A5000 GPU with 24GB VRAM.

\subsection{Comparison}
\label{sec:comparison}
We compare \MethodName{} with other task-driven ISP optimization methods on various downstream tasks, including object detection, instance segmentation, and image enhancement.
For each downstream task, \MethodName{} and competing methods are trained to find an optimal ISP pipeline that maximizes task performance. 
After training, the learned pipelines are applied to process test images, and the task performance is evaluated on the resulting outputs.

We benchmark against state-of-the-art approaches, including DRL-ISP~\cite{shin2022drlisp}, ReconfigISP~\cite{yu2021reconfigisp}, and AdaptiveISP~\cite{wang2024adaptiveisp}.
These methods differ in their input/output configurations: AdaptiveISP and our framework operate on RAW images that have undergone basic preprocessing operations, whereas DRL-ISP takes a Bayer input and produces a Bayer output without demosaicking. 
ReconfigISP, in contrast, incorporates demosaicking as part of its candidate modules, taking a Bayer RAW image and producing an sRGB output. 
To ensure a fair comparison under a consistent setting, we adapted both ReconfigISP and DRL-ISP to accept RAW images processed by the same preprocessing operations as ours. 
Further implementation details are provided in the supplementary material.

In addition, we include the in-camera ISP as a baseline for comparison on object detection and instance segmentation tasks.
The datasets we use provide both RAW and JPEG images, where the JPEGs are generated by in-camera ISPs embedded in commercial devices. 
This allows us to directly assess the performance of real camera ISPs alongside learned task-driven ISPs, highlighting how our method compares not only with prior research but also with the ISPs deployed in practice.

To maintain a fair comparison, we also align the training objectives across methods.
RL-based methods such as AdaptiveISP and DRL-ISP use rewards derived from $\mathcal{L}_\mathcal{T}$, while ReconfigISP directly optimizes $\mathcal{L}_\mathcal{T}$ under its neural architecture search formulation.

\vspace{-15pt}

\begin{figure*}[t]
    \centering
    \includegraphics[width=0.9\linewidth]{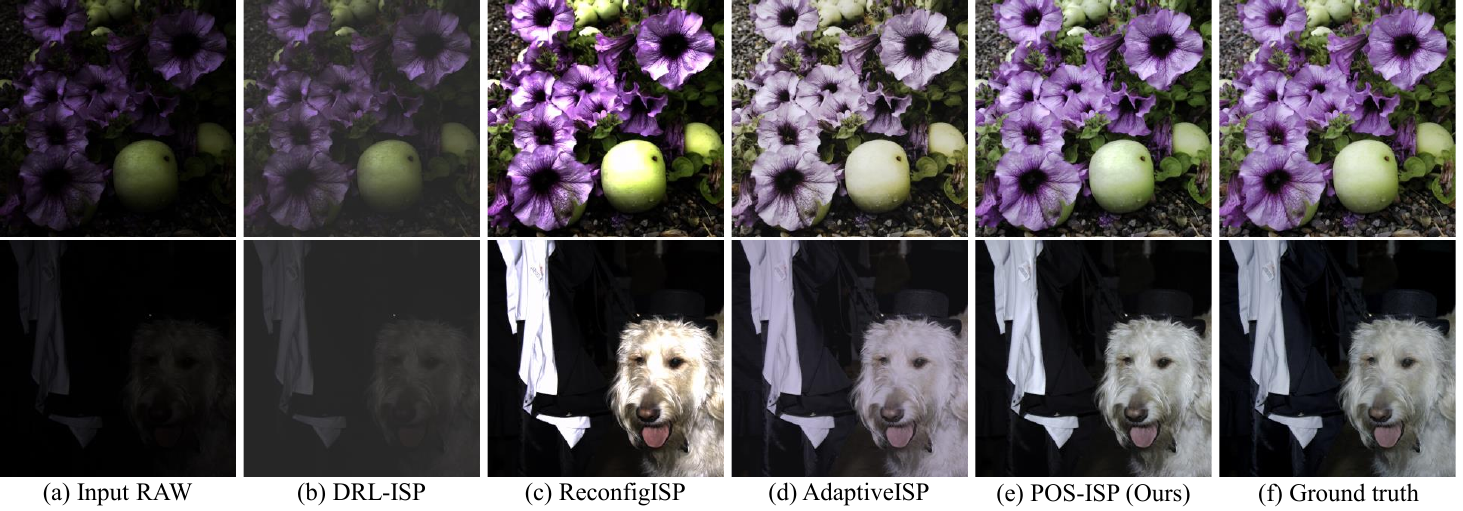}
    \vspace{-0.2cm}
\caption{\textbf{Qualitative comparison on image enhancement.} 
We use the images retouched by Expert C from the Adobe FiveK dataset as ground truth. Our method more closely matches the brightness and color tones of the ground truth.}
    \label{fig:comp_iqa}
    \vspace{-0.3cm}
\end{figure*}
\begin{table}[t]
\centering
\resizebox{0.95\columnwidth}{!}{
\begin{tabular}{l|ccc|ccc}
\toprule[1pt]
& \multicolumn{3}{c|}{\textbf{LOD-Dark}} 
& \multicolumn{3}{c}{\textbf{LOD-All}} \\
\cmidrule(lr){2-4}\cmidrule(lr){5-7}
\multirow{-3}{*}{\textbf{Method}} 
& \shortstack{mAP\\@0.5:0.95} & \shortstack{mAP\\@0.5} & \shortstack{mAP\\@0.75} 
& \shortstack{mAP\\@0.5:0.95} & \shortstack{mAP\\@0.5} & \shortstack{mAP\\@0.75}  \\
\midrule
Input RAW             & 44.1 & 67.7 & 47.5 & 53.6 & 70.5 & 57.5  \\
Camera ISP            & 37.6 & 55.4 & 41.6 & 48.8 & 64.5 & 52.2 \\
DRL-ISP~\cite{shin2022drlisp}        & 44.2 & 67.8 & 48.4 & 52.8 & 69.9 & 56.7 \\
ReconfigISP~\cite{yu2021reconfigisp} & 43.7 & 66.7 & 47.8 & 51.1 & 68.5 & 54.8 \\
AdaptiveISP~\cite{wang2024adaptiveisp} & 47.2 & 71.4 & 51.7 & 56.1 & 72.8 & 60.6 \\
\textbf{Ours}         & \textbf{47.8} & \textbf{72.1} & \textbf{52.8} & \textbf{56.6} & \textbf{73.1} & \textbf{60.9}  \\
\bottomrule[1pt]
\end{tabular}
}
\vspace{-0.2cm}
\caption{\textbf{Quantitative comparison with object detection task.} We highlight the best metrics as bold.}
\vspace{-0.4cm}
\label{tab:comp_detect}
\end{table}
\label{sec:comparison}
\paragraph{Object detection}
We first evaluate the effectiveness on the object detection task. 
Following AdaptiveISP~\cite{wang2024adaptiveisp}, we define the task loss $\mathcal{L}_\mathcal{T}$ as the sum of bounding box regression, objectness, and classification errors computed by the YOLOv3~\cite{redmon2018yolov3} detector pretrained on the COCO dataset~\cite{lin2014coco}, {with the detector parameters kept frozen during optimization.}
All methods are evaluated using the same detector to ensure consistency in the task loss definition. 

For training and evaluation, we employ the LOD dataset~\cite{Hong2021Crafting}, a real-world benchmark for low-light object detection. 
The dataset provides two subsets: LOD-Normal (well-lit) and LOD-Dark (low-light), each in JPEG and demosaicked RAW formats.  Following AdaptiveISP, we primarily use LOD-Dark to evaluate the performance gains of task-driven ISP optimization methods, including ours, under challenging low-light conditions. In addition, we use LOD-All, which combines LOD-Normal and LOD-Dark, to assess robustness to images with varying illuminations.

\cref{tab:comp_detect} and \cref{fig:comp_det_seg} present the quantitative and qualitative results, respectively.
In the table, ``Input RAW'' denotes images processed with the preprocessing operations assumed in our framework.
The results show that the in-camera ISP performs worse than the input RAW images, underscoring its limitations for task-driven objectives. 
Task-driven ISP methods generally outperform the in-camera ISP, but ReconfigISP suffers from a mismatch between soft training and hard inference. 
RL-based methods surpass the in-camera ISP yet remain suboptimal due to unstable reward estimation and stepwise formulation. 
In contrast, \MethodName{} achieves the highest accuracy by leveraging stable sequence-level optimization with accurate final rewards.

\begin{table}[t]
\centering
\resizebox{0.95\columnwidth}{!}{
\begin{tabular}{l|ccc|ccc}
\toprule[1pt]
& \multicolumn{3}{c|}{\textbf{LIS-Dark}} 
& \multicolumn{3}{c}{\textbf{LIS-All}} \\
\cmidrule(lr){2-4}\cmidrule(lr){5-7}
\multirow{-3}{*}{\textbf{Method}} 
& \shortstack{mAP\\@0.5:0.95} & \shortstack{mAP\\@0.5} & \shortstack{mAP\\@0.75}
& \shortstack{mAP\\@0.5:0.95} & \shortstack{mAP\\@0.5} & \shortstack{mAP\\@0.75} \\
\midrule
Input RAW            & 27.8 & 45.6 & 27.9 & 32.6 & 52.3 & 33.0 \\
Camera ISP           & 20.1 & 35.1 & 20.0 & 30.4 & 48.9 & 31.0 \\
DRL-ISP~\cite{shin2022drlisp} & 27.1 & 44.7 & 27.4 & 23.6 & 40.1 & 23.8 \\
ReconfigISP~\cite{yu2021reconfigisp} & 24.2 & 40.8 & 24.5 & 31.1 & 51.2 & 31.0 \\
AdaptiveISP~\cite{wang2024adaptiveisp} & 25.2 & 42.3 & 25.2 & 32.4 & 52.3 & 32.5 \\
\textbf{Ours}         & \textbf{32.1} & \textbf{51.8} & \textbf{32.1} & \textbf{34.9} & \textbf{55.9} & \textbf{34.9} \\
\bottomrule[1pt]
\end{tabular}
}
\vspace{-0.2cm}
\caption{\textbf{Quantitative comparison with instance segmentation task.} We highlight the best metrics as bold.} 
\vspace{-0.4cm}
\label{tab:comp_seg}
\end{table}

\vspace{-15pt}
\paragraph{Instance segmentation}
We further evaluate \MethodName{} on instance segmentation to assess its generalization to fine-grained vision tasks beyond object detection.
Here, $\mathcal{L}_\mathcal{T}$ is the sum of detection and mask losses from a YOLOv11-seg~\cite{khanam2024yolov11} model pretrained on the COCO dataset, {with the segmentation model parameters kept frozen during optimization.}
Evaluation is conducted on the LIS dataset~\cite{2023lis}, which consists of well-lit images (LIS-Normal) and low-light images (LIS-Dark). 
Similar to the object detection comparison, we consider LIS-Dark and LIS-All in this comparison, where LIS-All is the union of LIS-Dark and LIS-Normal. 

\cref{tab:comp_seg} and \cref{fig:comp_det_seg} present quantitative and qualitative comparisons, respectively.
Unlike object detection, task-driven ISP methods outperform the in-camera ISP but remain inferior to the input RAW images.
This highlights the difficulty of optimizing ISP pipelines for dense prediction tasks, where pixel-level supervision produces complex and unstable training signals.
In segmentation, small pixel deviations can disproportionately affect the reward, increasing variance and destabilizing learning.
For RL-based approaches, such high-variance rewards complicate value estimation and can lead to error accumulation during updates.
Our method avoids this issue by optimizing directly with task-level supervision rather than unstable value estimates, resulting in more stable training and better performance.

\vspace{-10pt}
\paragraph{Image enhancement}
{Lastly, we evaluate \MethodName{} on image enhancement using the Adobe FiveK dataset~\cite{fivek}. We adopt Expert C retouched images as the target style and define the task loss $\mathcal{L}_\mathcal{T}$ as the Mean Squared Error (MSE) between the ISP output and the corresponding Expert C image.}
Qualitative results are presented in \cref{fig:comp_iqa}.
{
DRL-ISP brightens the image but leaves many regions underexposed.
ReconfigISP improves brightness but produces overly saturated tones that deviate from the expert style.
AdaptiveISP shows noticeable color and white-balance shifts, producing desaturated tones.
In contrast, \MethodName{} produces visually pleasing results that closely match the desired retouching style.
}
These results demonstrate the robustness of our approach and its ability to generalize beyond recognition tasks to perceptual quality enhancement. Additional results are provided in the supplementary material.

\subsection{{Optimization Stability}}
\begin{figure}[t]
    \includegraphics[width=\linewidth]{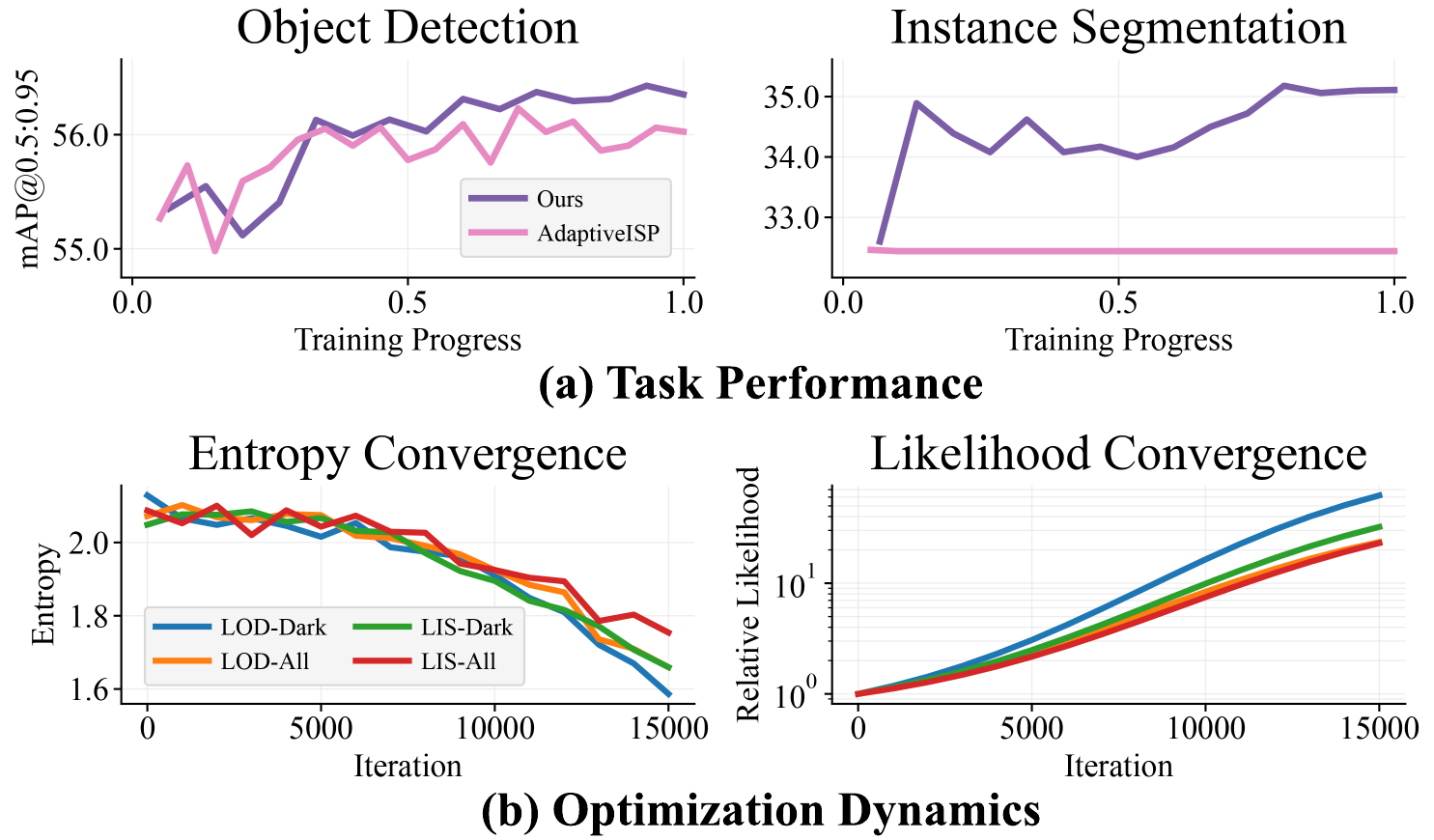}
\vspace{-8pt}
\caption{
\textbf{Optimization behavior.} (a) Task score on the test set over training progress.
(b) (left) policy entropy convergence and (right) relative likelihood of the final pipeline.
}
\label{fig:stats}
\vspace{-0.3cm}
\end{figure}

\paragraph{Training dynamics}
We further analyze the optimization behavior during training.
In \cref{fig:stats}-(a), we plot the test performance curves for object detection and instance segmentation on the LOD-All~\cite{Hong2021Crafting} and LIS-All~\cite{2023lis}, respectively, against the training progress, defined as the ratio between the checkpoint iteration and the total training iterations.
\MethodName{} improves steadily throughout training, whereas AdaptiveISP~\cite{wang2024adaptiveisp} either exhibits noticeable fluctuations or improves only marginally early in training, suggesting that our method shows more stable optimization behavior.

The optimization statistics in \cref{fig:stats}-(b) further support this observation.
The policy entropy steadily decreases during training, suggesting that the policy becomes increasingly confident in selecting pipelines. 
At the same time, the likelihood assigned to the pipeline selected by the final policy consistently increases, growing by approximately \(20\text{--}60\times\) across datasets.
This likelihood is computed by summing the log probabilities of the selected modules to obtain the pipeline log-likelihood and exponentiating its difference from the initial log-likelihood.
Together, these trends indicate that the policy progressively concentrates its probability mass on effective pipelines, resulting in stable optimization.
\begin{table}[t]
\centering
\resizebox{\columnwidth}{!}{
\begin{tabular}{lccc}
\toprule
\textbf{Method} 
& {mAP @0.5:0.95} 
& {mAP @0.5} 
& {mAP @0.75} \\
\midrule
DRL-ISP~\cite{shin2022drlisp}
& 44.00 $\pm$ 0.20 
& 67.73 $\pm$ 0.47 
& 47.77 $\pm$ 0.31 \\

AdaptiveISP~\cite{wang2024adaptiveisp}
& 47.13 $\pm$ 0.25 
& 71.17 $\pm$ 0.15 
& 51.73 $\pm$ 0.57 \\

\textbf{Ours} 
& \textbf{47.80 $\pm$\ 0.08} 
& \textbf{72.10 $\pm$\ 0.16} 
& \textbf{52.67 $\pm$\ 0.12} \\
\bottomrule
\end{tabular}}
\vspace{-0.2cm}
\caption{\textbf{Multi-seed quantitative comparison on the LOD-Dark object detection benchmark.}
We report mean $\pm$ standard deviation over three seeds.}
\vspace{-0.1cm}
\label{tab:detection_multiseed_lod}
\end{table}

\vspace{-15pt}
\paragraph{Multi-seed comparison}
We examine the robustness of the optimization across random seeds. On the LOD-Dark~\cite{Hong2021Crafting} object detection benchmark with a YOLOv3~\cite{redmon2018yolov3} detector, the three-seed results (\cref{tab:detection_multiseed_lod}) show that \MethodName{} consistently outperforms DRL-ISP~\cite{shin2022drlisp} and AdaptiveISP~\cite{wang2024adaptiveisp}, while exhibiting substantially smaller standard deviations.
This indicates that our method achieves reproducible gains and consistently attains high performance across independent training runs.


\begin{table}[t]
\centering
\resizebox{\columnwidth}{!}{
\begin{tabular}{l|cccc}
\toprule
Method                                 & Params (M)    & MACs (M)      & Peak GPU Memory (MB) & Runtime (ms) \\
\midrule
DRL-ISP~\cite{shin2022drlisp}     & 6.57          & 155.3         & 1013.9               & 15.71 \\
AdaptiveISP~\cite{wang2024adaptiveisp} & 7.18          & 70.2          & 39.6                 & 12.72 \\
\textbf{Ours}                                   & \textbf{0.53} & \textbf{15.1} & \textbf{14.4}        & \textbf{1.55} \\
\bottomrule
\end{tabular}}
\vspace{-0.2cm}
\caption{\textbf{Comparison of computational efficiency.} All results are measured on a single NVIDIA RTX 2080 Ti with inputs of resolution $512\times512$. We excluded the module execution time when measuring the runtime.  We highlight the best metrics as bold.}
\vspace{-0.3cm}
\label{tab:efficiency}
\end{table}

\subsection{Computational Efficiency}
We compare the inference efficiency of \MethodName{} with RL-based methods~\cite{shin2022drlisp, wang2024adaptiveisp}.
For DRL-ISP~\cite{shin2022drlisp} and AdaptiveISP~\cite{wang2024adaptiveisp}, runtime is measured using pipelines of three and five modules, respectively, following the original settings.
The results are summarized in \cref{tab:efficiency}.
{RL-based approaches incur considerable computational overhead because the controller must be executed repeatedly during pipeline construction. DRL-ISP further amplifies this cost by employing a heavy feature extractor, resulting in substantial computational and memory demands. AdaptiveISP reduces memory usage compared to DRL-ISP, but still relies on a relatively large controller and repeated inference, leading to non-negligible overhead.
In contrast, \MethodName{} is designed to be lightweight and efficient. By fixing the module sequence at inference time, it predicts only the module parameters in a single forward pass, eliminating repeated controller execution and iterative decision-making. This significantly reduces both computational cost and memory usage, enabling superior efficiency with minimal overhead.}

\subsection{Ablation on Sequence Predictor}
Our sequence predictor is designed to capture inter-module dependencies in ISP pipeline prediction. 
To evaluate its impact, we conduct an ablation study by constructing a variant in which \ActionNetName{} is replaced with a learnable probability table. 
In this table, the element at row $i$ and column $j$ denotes the probability of selecting module $a_j$ at step $i$, thereby modeling each decision independently without considering contextual relationships. 
This design allows a direct examination of the benefits of exploiting inter-module dependencies when predicting module sequences. 
We use the same experimental settings described in \cref{sec:comparison}.

As shown in \cref{tab:ablation_dependency}, the probability-table variant (\cref{tab:ablation_dependency}-(a)) fails to capture contextual relationships among modules, resulting in lower performance. 
By contrast, adopting a recurrent structure (\cref{tab:ablation_dependency}-(b)) enables the model to learn inter-module dependencies, leading to improved performance by effectively modeling the influence of module order on sequence prediction.

\begin{table}[t]
\centering
\resizebox{\columnwidth}{!}{
\begin{tabular}{l|ccc|ccc}
\toprule
\multicolumn{1}{c|}{\multirow{3}{*}{\shortstack{Sequence\\predictor}}} &
\multicolumn{3}{c|}{\textbf{LOD-Dark}} &
\multicolumn{3}{c}{\textbf{LIS-Dark}} \\
\cmidrule(lr){2-4} \cmidrule(lr){5-7}
& mAP       & mAP  & mAP   & mAP       & mAP  & mAP \\
& @0.5:0.95 & @0.5 & @0.75 & @0.5:0.95 & @0.5 & @0.75 \\
\midrule
(a) Prob. table & 47.5&71.4&52.1&31.3&50.9&31.9  \\
(b) GRU (Ours) & \textbf{47.8} & \textbf{72.1} & \textbf{52.8} & \textbf{32.1} & \textbf{51.8} & \textbf{32.1} \\
\bottomrule
\end{tabular}
}
\vspace{-0.2cm}
\caption{
\textbf{Ablation study on the effect of adding recurrent sequence estimation.} We highlight the best metrics as bold.
}
\vspace{-0.3cm}
\label{tab:ablation_dependency}
\end{table}

\section{Conclusion}
\label{sec:conclustion}

In this paper, we introduce \MethodName{}, a task-aware ISP optimization framework that jointly learns module sequences and parameters for a target downstream task.
By eliminating the unstable future reward estimation and stepwise decision process, our approach enables stable optimization while explicitly modeling inter-module dependencies.
\MethodName{} achieves state-of-the-art accuracy on object detection and instance segmentation with substantially lower computational and memory overhead, and also shows promising results on image enhancement.
\vspace{-5pt}

\paragraph{Limitations and future work}
Despite promising results, our framework still has some limitations. Increasing the number of ISP candidate modules enlarges the search space and may require longer training to achieve convergence. Moreover, the current system requires separate training for each downstream task, which limits scalability in multi-task settings, as different tasks rely on independently optimized ISP policies. This increases system complexity when multiple tasks need to be supported. As future work, extending the framework to a unified model that can jointly optimize multiple tasks could improve scalability and broaden its applicability.

\paragraph{Acknowledgement} We thank Junpyo Seo for his assistance with the on-device deployment test. This work was supported by Samsung Electronics Co., Ltd (IO251210-14286-01), and by the Institute of Information \& Communications Technology Planning \& Evaluation (IITP) grant funded by the Korea government (MSIT) (IITP-2026-RS-2024-00437866, No.RS-2019-II191906, Artificial Intelligence Graduate School Program (POSTECH)).

{
    \small
    \bibliographystyle{ieeenat_fullname}
    \bibliography{main}

\begin{thebibliography}{28}
\providecommand{\natexlab}[1]{#1}
\providecommand{\url}[1]{\texttt{#1}}
\expandafter\ifx\csname urlstyle\endcsname\relax
  \providecommand{\doi}[1]{doi: #1}\else
  \providecommand{\doi}{doi: \begingroup \urlstyle{rm}\Url}\fi

\bibitem[Afifi et~al.(2021)Afifi, Abdelhamed, Abuolaim, Punnappurath, and Brown]{afifi2021ciexyznet}
Mahmoud Afifi, Abdelrahman Abdelhamed, Abdullah Abuolaim, Abhijith Punnappurath, and Michael~S Brown.
\newblock Cie xyz net: Unprocessing images for low-level computer vision tasks.
\newblock \emph{IEEE TPAMI}, 2021.

\bibitem[Brooks et~al.(2019)Brooks, Mildenhall, Xue, Chen, Sharlet, and Barron]{brooks2019upi}
Tim Brooks, Ben Mildenhall, Tianfan Xue, Jiawen Chen, Dillon Sharlet, and Jonathan~T Barron.
\newblock Unprocessing images for learned raw denoising.
\newblock In \emph{CVPR}, 2019.

\bibitem[Bychkovsky et~al.(2011)Bychkovsky, Paris, Chan, and Durand]{fivek}
Vladimir Bychkovsky, Sylvain Paris, Eric Chan, and Fr{\'e}do Durand.
\newblock Learning photographic global tonal adjustment with a database of input / output image pairs.
\newblock In \emph{CVPR}, 2011.

\bibitem[Cesa-Bianchi et~al.(2017)Cesa-Bianchi, Gentile, Lugosi, and Neu]{cesa2017boltzmann}
Nicol{\`o} Cesa-Bianchi, Claudio Gentile, G{\'a}bor Lugosi, and Gergely Neu.
\newblock Boltzmann exploration done right.
\newblock In \emph{NeurIPS}, 2017.

\bibitem[Chen et~al.(2018)Chen, Chen, Xu, and Koltun]{chen2018learning}
Chen Chen, Qifeng Chen, Jia Xu, and Vladlen Koltun.
\newblock Learning to see in the dark.
\newblock In \emph{CVPR}, 2018.

\bibitem[Chen et~al.(2025)Chen, Xiao, Zhang, Shi, and Gu]{chen2025task}
Kai Chen, Jin Xiao, Leheng Zhang, Kexuan Shi, and Shuhang Gu.
\newblock Task-aware image signal processor for advanced visual perception.
\newblock \emph{arXiv}, 2025.

\bibitem[Chen et~al.(2023)Chen, Fu, Wei, Zheng, and Heide]{2023lis}
Linwei Chen, Ying Fu, Kaixuan Wei, Dezhi Zheng, and Felix Heide.
\newblock Instance segmentation in the dark.
\newblock \emph{IJCV}, 2023.

\bibitem[Cho et~al.(2014)Cho, Van~Merri{\"e}nboer, Bahdanau, and Bengio]{cho2014gru}
Kyunghyun Cho, Bart Van~Merri{\"e}nboer, Dzmitry Bahdanau, and Yoshua Bengio.
\newblock On the properties of neural machine translation: Encoder-decoder approaches.
\newblock \emph{arXiv}, 2014.

\bibitem[Conde et~al.(2022)Conde, McDonagh, Maggioni, Leonardis, and P{\'e}rez-Pellitero]{conde2022model}
Marcos~V Conde, Steven McDonagh, Matteo Maggioni, Ales Leonardis, and Eduardo P{\'e}rez-Pellitero.
\newblock Model-based image signal processors via learnable dictionaries.
\newblock In \emph{AAAI}, 2022.

\bibitem[Fujimoto et~al.(2018)Fujimoto, van Hoof, and Meger]{fujimoto2018addressing}
Scott Fujimoto, Herke van Hoof, and David Meger.
\newblock Addressing function approximation error in actor-critic methods.
\newblock In \emph{ICML}, 2018.

\bibitem[Hidasi et~al.(2015)Hidasi, Karatzoglou, Baltrunas, and Tikk]{hidasi2015session}
Bal{\'a}zs Hidasi, Alexandros Karatzoglou, Linas Baltrunas, and Domonkos Tikk.
\newblock Session-based recommendations with recurrent neural networks.
\newblock \emph{arXiv}, 2015.

\bibitem[Hong et~al.(2021)Hong, Wei, Chen, and Fu]{Hong2021Crafting}
Yang Hong, Kaixuan Wei, Linwei Chen, and Ying Fu.
\newblock Crafting object detection in very low light.
\newblock In \emph{BMVC}, 2021.

\bibitem[Khanam and Hussain(2024)]{khanam2024yolov11}
Rahima Khanam and Muhammad Hussain.
\newblock Yolov11: An overview of the key architectural enhancements.
\newblock \emph{arXiv}, 2024.

\bibitem[Kim et~al.(2024)Kim, Kim, Lee, Lee, Baek, and Cho]{kim2024paramisp}
Woohyeok Kim, Geonu Kim, Junyong Lee, Seungyong Lee, Seung-Hwan Baek, and Sunghyun Cho.
\newblock Paramisp: Learned forward and inverse isps using camera parameters.
\newblock In \emph{CVPR}, 2024.

\bibitem[Kingma and Ba(2014)]{kingma2014adam}
Diederik~P. Kingma and Jimmy Ba.
\newblock Adam: A method for stochastic optimization.
\newblock \emph{arXiv}, 2014.

\bibitem[Lin et~al.(2014)Lin, Maire, Belongie, Hays, Perona, Ramanan, Doll{\'a}r, and Zitnick]{lin2014coco}
Tsung-Yi Lin, Michael Maire, Serge Belongie, James Hays, Pietro Perona, Deva Ramanan, Piotr Doll{\'a}r, and C.~Lawrence Zitnick.
\newblock Microsoft coco: Common objects in context.
\newblock In \emph{ECCV}, 2014.

\bibitem[Qin et~al.(2022)Qin, Han, Wang, Zhang, Li, Li, and Hu]{qin2022attention}
Haina Qin, Longfei Han, Juan Wang, Congxuan Zhang, Yanwei Li, Bing Li, and Weiming Hu.
\newblock Attention-aware learning for hyperparameter prediction in image processing pipelines.
\newblock In \emph{ECCV}, 2022.

\bibitem[Redmon and Farhadi(2018)]{redmon2018yolov3}
Joseph Redmon and Ali Farhadi.
\newblock Yolov3: An incremental improvement.
\newblock \emph{arXiv}, 2018.

\bibitem[Ren et~al.(2019)Ren, Chen, Li, Ren, Ma, and de~Rijke]{ren2019repeatnet}
Pengjie Ren, Zhumin Chen, Jing Li, Zhaochun Ren, Jun Ma, and Maarten de Rijke.
\newblock Repeatnet: A repeat-aware neural recommendation machine for session-based recommendation.
\newblock In \emph{AAAI}, 2019.

\bibitem[Serban et~al.(2016)Serban, Sordoni, Bengio, Courville, and Pineau]{serban2016building}
Iulian Serban, Alessandro Sordoni, Yoshua Bengio, Aaron Courville, and Joelle Pineau.
\newblock Building end-to-end dialogue systems using generative hierarchical neural network models.
\newblock In \emph{AAAI}, 2016.

\bibitem[Shin et~al.(2022)Shin, Lee, and Kweon]{shin2022drlisp}
Ukcheol Shin, Kyunghyun Lee, and In~So Kweon.
\newblock Drl-isp: Multi-objective camera isp with deep reinforcement learning.
\newblock In \emph{IROS}, 2022.

\bibitem[Tseng et~al.(2019)Tseng, Yu, Yang, Mannan, Arnaud, Nowrouzezahrai, Lalonde, and Heide]{tseng2019hyperparameter}
Ethan Tseng, Felix Yu, Yuting Yang, Fahim Mannan, Karl~ST Arnaud, Derek Nowrouzezahrai, Jean-Fran{\c{c}}ois Lalonde, and Felix Heide.
\newblock Hyperparameter optimization in black-box image processing using differentiable proxies.
\newblock \emph{ACM TOG}, 2019.

\bibitem[van Hasselt et~al.(2018)van Hasselt, Doron, Strub, Hessel, Sonnerat, and Modayil]{van2018deep}
Hado van Hasselt, Yotam Doron, Florian Strub, Matteo Hessel, Nicolas Sonnerat, and Joseph Modayil.
\newblock Deep reinforcement learning and the deadly triad.
\newblock \emph{arXiv}, 2018.

\bibitem[Wang et~al.(2024)Wang, Xu, Zhang, Xue, and Gu]{wang2024adaptiveisp}
Yujin Wang, Tianyi Xu, Fan Zhang, Tianfan Xue, and Jinwei Gu.
\newblock Adaptiveisp: Learning an adaptive image signal processor for object detection.
\newblock In \emph{NeurIPS}, 2024.

\bibitem[Williams(1992)]{williams1992simple}
Ronald~J. Williams.
\newblock Simple statistical gradient-following algorithms for connectionist reinforcement learning.
\newblock \emph{Machine Learning}, 1992.

\bibitem[Xing et~al.(2021)Xing, Qian, and Chen]{xing2021invisp}
Yazhou Xing, Zian Qian, and Qifeng Chen.
\newblock Invertible image signal processing.
\newblock In \emph{CVPR}, 2021.

\bibitem[Yu et~al.(2021)Yu, Li, Peng, Loy, and Gu]{yu2021reconfigisp}
Ke Yu, Zexian Li, Yue Peng, Chen~Change Loy, and Jinwei Gu.
\newblock Reconfigisp: Reconfigurable camera image processing pipeline.
\newblock In \emph{ICCV}, 2021.

\bibitem[Zamir et~al.(2020)Zamir, Arora, Khan, Hayat, Khan, Yang, and Shao]{zamir2020cycleisp}
Syed~Waqas Zamir, Aditya Arora, Salman Khan, Munawar Hayat, Fahad~Shahbaz Khan, Ming-Hsuan Yang, and Ling Shao.
\newblock Cycleisp: Real image restoration via improved data synthesis.
\newblock In \emph{CVPR}, 2020.

\end{thebibliography}


\begin{thebibliography}{15}
\providecommand{\natexlab}[1]{#1}
\providecommand{\url}[1]{\texttt{#1}}
\expandafter\ifx\csname urlstyle\endcsname\relax
  \providecommand{\doi}[1]{doi: #1}\else
  \providecommand{\doi}{doi: \begingroup \urlstyle{rm}\Url}\fi

\bibitem[Bian et~al.(2019)Bian, Li, Wang, Zhan, Shen, Cheng, and Reid]{bian2019unsupervised}
Jiawang Bian, Zhichao Li, Naiyan Wang, Huangying Zhan, Chunhua Shen, Ming-Ming Cheng, and Ian Reid.
\newblock Unsupervised scale-consistent depth and ego-motion learning from monocular video.
\newblock In \emph{NeurIPS}, 2019.

\bibitem[Bychkovsky et~al.(2011)Bychkovsky, Paris, Chan, and Durand]{fivek}
Vladimir Bychkovsky, Sylvain Paris, Eric Chan, and Fr{\'e}do Durand.
\newblock Learning photographic global tonal adjustment with a database of input / output image pairs.
\newblock In \emph{CVPR}, 2011.

\bibitem[Chen et~al.(2023)Chen, Fu, Wei, Zheng, and Heide]{2023lis}
Linwei Chen, Ying Fu, Kaixuan Wei, Dezhi Zheng, and Felix Heide.
\newblock Instance segmentation in the dark.
\newblock \emph{IJCV}, 2023.

\bibitem[Geiger et~al.(2012)Geiger, Lenz, and Urtasun]{geiger2012we}
Andreas Geiger, Philip Lenz, and Raquel Urtasun.
\newblock Are we ready for autonomous driving? the kitti vision benchmark suite.
\newblock In \emph{CVPR}, 2012.

\bibitem[Hong et~al.(2021)Hong, Wei, Chen, and Fu]{Hong2021Crafting}
Yang Hong, Kaixuan Wei, Linwei Chen, and Ying Fu.
\newblock Crafting object detection in very low light.
\newblock In \emph{BMVC}, 2021.

\bibitem[Hu et~al.(2018)Hu, He, Xu, Wang, and Lin]{hu2018exposure}
Yuanming Hu, Hao He, Chenxi Xu, Baoyuan Wang, and Stephen Lin.
\newblock Exposure: A white-box photo post-processing framework.
\newblock \emph{ACM TOG}, 2018.

\bibitem[Jelley et~al.(2024)Jelley, McInroe, Devlin, and Storkey]{jelley2024efficient}
Adam Jelley, Trevor McInroe, Sam Devlin, and Amos Storkey.
\newblock Efficient offline reinforcement learning: The critic is critical.
\newblock \emph{arXiv}, 2024.

\bibitem[Lei et~al.(2025)Lei, Li, Wu, Hu, Zhou, Zheng, Ding, Du, Wu, and Gao]{lei2025yolov13}
Mengqi Lei, Siqi Li, Yihong Wu, Han Hu, You Zhou, Xinhu Zheng, Guiguang Ding, Shaoyi Du, Zongze Wu, and Yue Gao.
\newblock Yolov13: Real-time object detection with hypergraph-enhanced adaptive visual perception.
\newblock \emph{arXiv}, 2025.

\bibitem[Metelli et~al.(2021)Metelli, Pirotta, Calandriello, and Restelli]{metelli2021safe}
Alberto~Maria Metelli, Matteo Pirotta, Daniele Calandriello, and Marcello Restelli.
\newblock Safe policy iteration: A monotonically improving approximate policy iteration approach.
\newblock \emph{JMLR}, 2021.

\bibitem[Perez et~al.(2017)Perez, Strub, de~Vries, Dumoulin, and Courville]{perez2017filmvisualreasoninggeneral}
Ethan Perez, Florian Strub, Harm de Vries, Vincent Dumoulin, and Aaron Courville.
\newblock Film: Visual reasoning with a general conditioning layer.
\newblock \emph{arXiv}, 2017.

\bibitem[Redmon and Farhadi(2018)]{redmon2018yolov3}
Joseph Redmon and Ali Farhadi.
\newblock Yolov3: An incremental improvement.
\newblock \emph{arXiv}, 2018.

\bibitem[Shin et~al.(2022)Shin, Lee, and Kweon]{shin2022drlisp}
Ukcheol Shin, Kyunghyun Lee, and In~So Kweon.
\newblock Drl-isp: Multi-objective camera isp with deep reinforcement learning.
\newblock In \emph{IROS}, 2022.

\bibitem[Wagner(2011)]{wagner2011reinterpretation}
Paul Wagner.
\newblock A reinterpretation of the policy oscillation phenomenon in approximate policy iteration.
\newblock In \emph{NeurIPS}, 2011.

\bibitem[Wang et~al.(2024)Wang, Xu, Zhang, Xue, and Gu]{wang2024adaptiveisp}
Yujin Wang, Tianyi Xu, Fan Zhang, Tianfan Xue, and Jinwei Gu.
\newblock Adaptiveisp: Learning an adaptive image signal processor for object detection.
\newblock In \emph{NeurIPS}, 2024.

\bibitem[Yu et~al.(2021)Yu, Li, Peng, Loy, and Gu]{yu2021reconfigisp}
Ke Yu, Zexian Li, Yue Peng, Chen~Change Loy, and Jinwei Gu.
\newblock Reconfigisp: Reconfigurable camera image processing pipeline.
\newblock In \emph{ICCV}, 2021.

\end{thebibliography}
}

\end{document}


\maketitle
\setcounter{page}{1}
\setcounter{section}{0}
\setcounter{table}{0}
\setcounter{figure}{0}

{
  \renewcommand\thesection{\arabic{section}}
  \renewcommand\thetable{\arabic{table}}
  \renewcommand\thefigure{\arabic{figure}}
  \tableofcontents
}

\renewcommand\thesection{S\arabic{section}}
\renewcommand\thetable{S\arabic{table}}
\renewcommand\thefigure{S\arabic{figure}}

\thispagestyle{empty}

\section{Additional Results}

\subsection{Additional Qualitative Results}
\begin{figure*}[t]
    \centering
    \includegraphics[width=\linewidth]{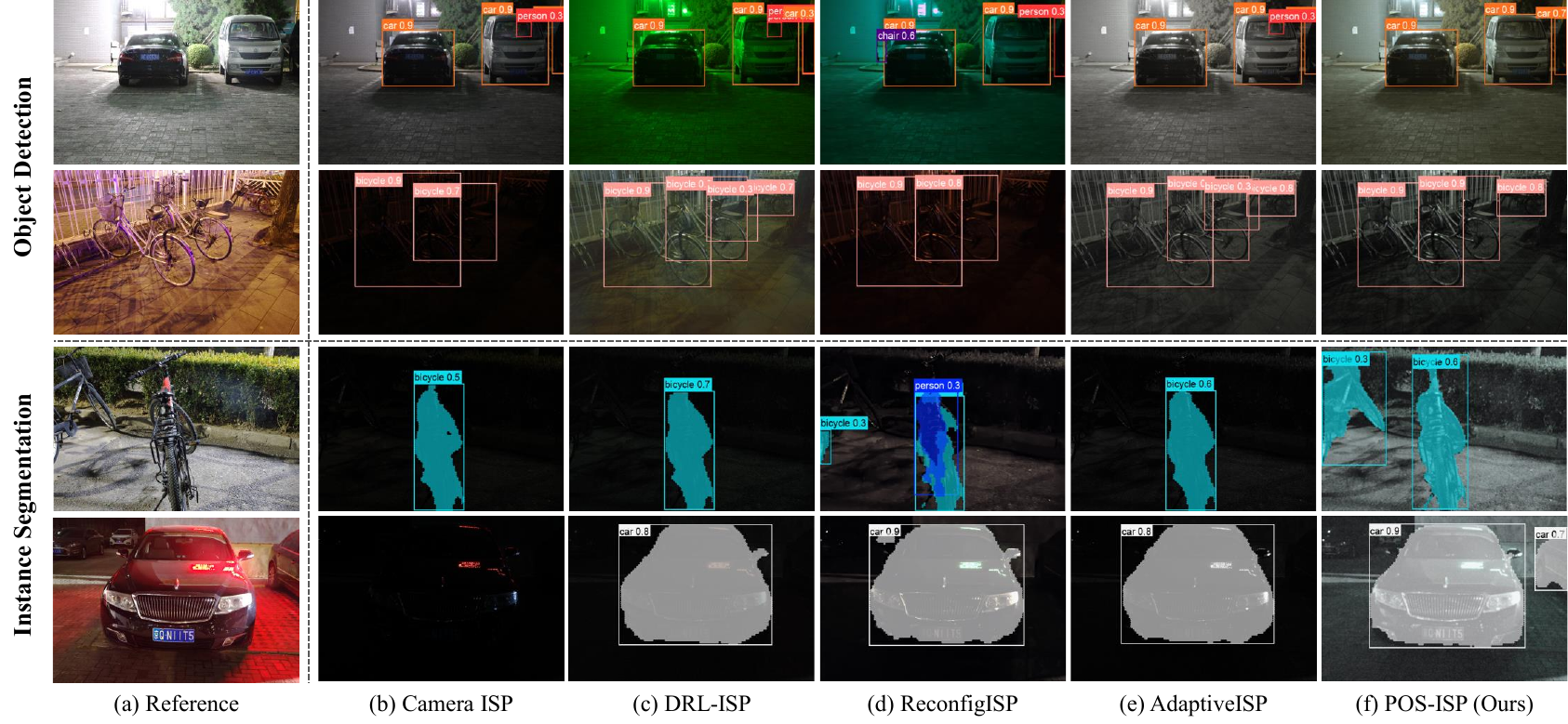}
    \caption{\textbf{Comparison of different ISP methods on object detection and instance segmentation tasks.} Reference images are from the LOD-Normal and LIS-Normal datasets with brightness increased by $1.5\times$ for visualization. 
    }
    \vspace{-0.3cm}
    \label{fig:comp_det_seg_supple}
\end{figure*}
\begin{figure*}[p]
    \centering
    \includegraphics[width=0.895\linewidth]{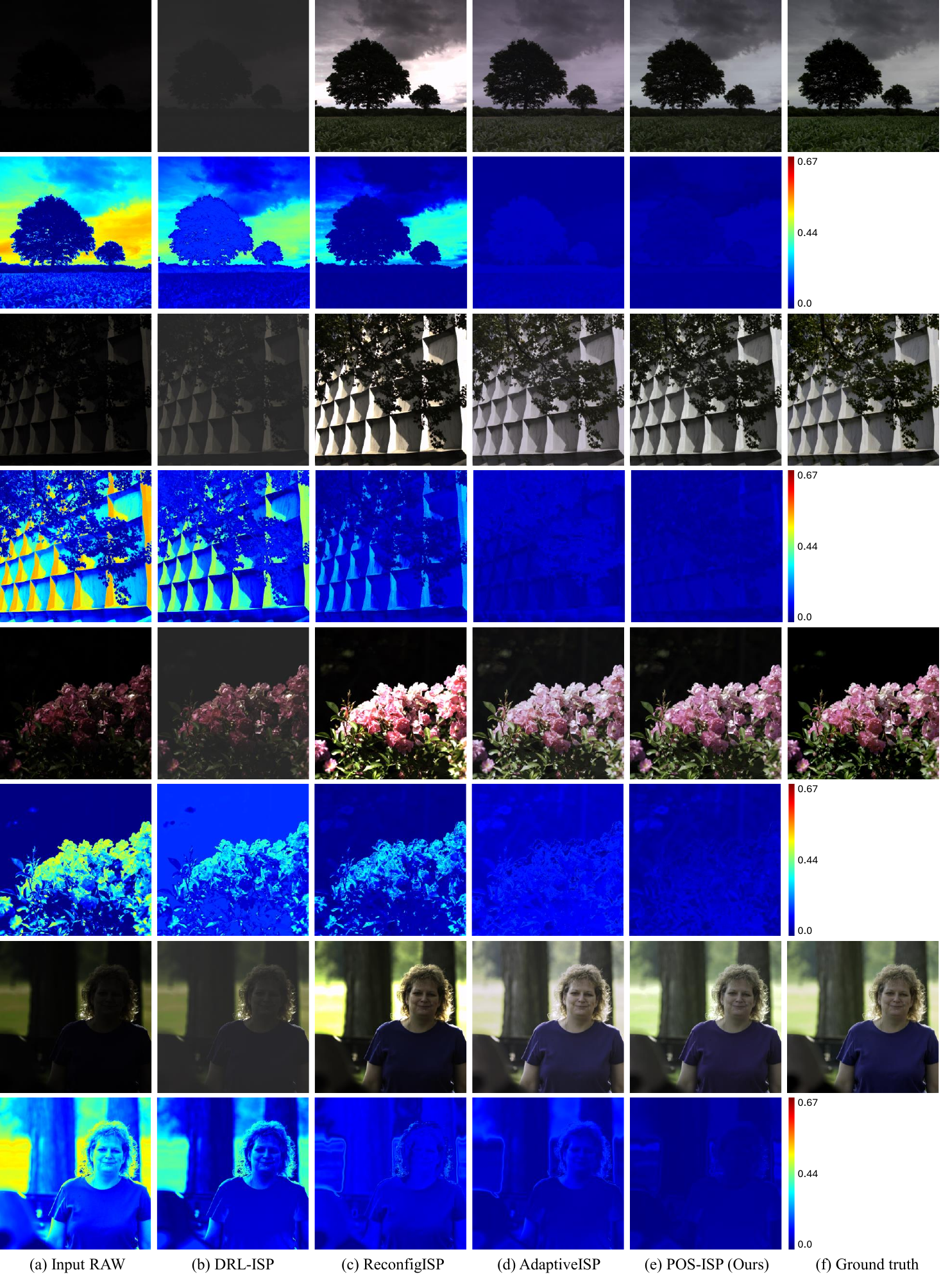}
    \caption{\textbf{Qualitative comparison on the image enhancement task.} 
    We use the images retouched by Expert C from the Adobe FiveK dataset as ground truth. 
    Error maps are also visualized to highlight pixel-wise differences from the ground truth.}
    \label{fig:iqa_comp_supple}
\end{figure*}

Additional qualitative examples for object detection and instance segmentation are shown in \cref{fig:comp_det_seg_supple}, and additional results for image enhancement are provided in \cref{fig:iqa_comp_supple}.
These results provide further visual references that complement the comparisons presented in the main paper.

\subsection{{Image Enhancement}}
We provide quantitative comparisons for the image enhancement task described in the main paper. 
The results are summarized in \cref{tab:comp_iqa}. Existing task-driven ISP optimization methods improve quantitative metrics compared to the input images, indicating that optimizing the ISP pipeline with task supervision can benefit perceptual enhancement. However, their performance remains limited, and the outputs often fail to fully reproduce the desired style.
DRL-ISP~\cite{shin2022drlisp} tends to uniformly increase image intensity due to unstable policy learning, leaving some regions under-enhanced. ReconfigISP~\cite{yu2021reconfigisp} applies a fixed pipeline to all images, which often produces overly strong contrast and tone deviations from the target style. AdaptiveISP~\cite{wang2024adaptiveisp} achieves relatively strong performance but fails to consistently find the optimal configurations, resulting in color and white balance shifts.
In contrast, \MethodName{} achieves the best performance across the reported metrics. This result suggests that our sequence-level optimization framework can effectively learn ISP configurations that align with the target retouching style.

\begin{table}[t]
\centering
\resizebox{0.8\columnwidth}{!}{
\begin{tabular}{l|ccc}
\toprule[1pt]
\textbf{Method} & PSNR $\uparrow$  & SSIM $\uparrow$  & LPIPS $\downarrow$  \\
\midrule
Input RAW                    & 9.18 & 0.195 & 0.585  \\
DRL-ISP~\cite{shin2022drlisp}          & 13.68 & 0.571 & 0.492 \\
ReconfigISP~\cite{yu2021reconfigisp}   & 17.28 & 0.818 & 0.216\\
AdaptiveISP~\cite{wang2024adaptiveisp} & 22.73	& 0.908	& 0.105 \\
\textbf{Ours}         & \textbf{23.11} & \textbf{0.923}& \textbf{0.097} \\
\bottomrule[1pt]
\end{tabular}
}
\vspace{-0.1cm}
\caption{\textbf{Quantitative comparison for full-reference image quality enhancement task.} In the table, ``Input RAW'' denotes images that have undergone only the most basic preprocessing steps using Adobe Lightroom. We highlight the best metrics as bold.}
\label{tab:comp_iqa}
\end{table}

\subsection{{Depth Estimation}}
\begin{table}[t]
\centering
\resizebox{0.98\columnwidth}{!}{
\begin{tabular}{l|cccc|c}
\toprule
\multirow{2}{*}{\textbf{Method}} 
& \multicolumn{4}{c|}{\textbf{Error} $\downarrow$} 
& \textbf{Accuracy} $\uparrow$ \\
\cmidrule(lr){2-5} \cmidrule(lr){6-6}
& AbsRel & SqRel & RMS & RMSlog & $\delta < 1.25$ \\
\midrule
Input RAW        & 0.293 & 2.560 & 8.73 & 0.398 & 0.493 \\
DRL-ISP~\cite{shin2022drlisp}           & 0.131 & 0.978 & 5.22 & 0.209 & 0.839 \\
ReconfigISP~\cite{yu2021reconfigisp}      & 0.154 & 1.128 & 5.76 & 0.235 & 0.788 \\
AdaptiveISP~\cite{wang2024adaptiveisp}      & 0.131 & 0.971 & 5.23 & 0.207 & 0.840 \\
\textbf{Ours}    & \textbf{0.128} & \textbf{0.947} & \textbf{5.13} & \textbf{0.205} & \textbf{0.847} \\
\bottomrule
\end{tabular}
}
\vspace{-0.1cm}
\caption{\textbf{Quantitative comparison with depth estimation task.} We highlight the best metrics as bold.} 
\vspace{-0.3cm}
\label{tab:depth_rms}
\end{table}

We further evaluate \MethodName{} on monocular depth estimation to examine its effectiveness in dense geometric prediction tasks. In this setting, the task loss $\mathcal{L}_{\mathcal{T}}$ is defined using the Root Mean Squared Error (RMSE) objective, which emphasizes large depth discrepancies and penalizes outliers. Following DRL-ISP~\cite{shin2022drlisp}, depth supervision is obtained from SC-SfMLearner~\cite{bian2019unsupervised} predictions on the KITTI~\cite{geiger2012we} dataset. 
{For fair comparison, the same task loss and depth predictor are consistently applied to all methods.}
Performance is evaluated using standard depth estimation metrics, including AbsRel, SqRel, RMS, RMS${\log}$, and accuracy ($\delta < 1.25$). 

\cref{tab:depth_rms} presents the results when models are trained using the RMSE objective. Compared to existing ISP optimization methods, \MethodName{} consistently achieves the best performance across all evaluation metrics. These improvements are observed not only in RMS but also in relative error metrics and accuracy, indicating that the gains are not restricted to a single evaluation criterion.

Overall, these results demonstrate that the performance gains of \MethodName{} are not confined to recognition-based tasks. 
Instead, the method generalizes effectively to dense geometric prediction, highlighting its robustness and effectiveness across diverse downstream tasks, including detection, segmentation, image enhancement, and depth estimation.

\section{Additional Analysis}
\subsection{{Optimization Stability}}
\begin{table}[t]
\centering
\resizebox{\columnwidth}{!}{
\begin{tabular}{l|c|ccc|ccc}
\toprule[1pt]
\textbf{Method} & \textbf{Dataset}
& \multicolumn{3}{c|}{\textbf{Object Detection (LOD)}}
& \multicolumn{3}{c}{\textbf{Instance Segmentation (LIS)}} \\
\cmidrule(lr){3-5} \cmidrule(lr){6-8}
& 
& \shortstack{mAP\\@0.5:0.95} & \shortstack{mAP\\@0.5} & \shortstack{mAP\\@0.75}
& \shortstack{mAP\\@0.5:0.95} & \shortstack{mAP\\@0.5} & \shortstack{mAP\\@0.75} \\
\midrule
AdaptiveISP & Dark
& 47.2 & 71.4 & 51.7
& 25.2 & 42.3 & 25.2 \\
\textbf{Ours} & Dark
& \textbf{47.8 \small{$\pm$ 0.08}} & \textbf{72.1 \small{$\pm$ 0.16}} & \textbf{52.7 \small{$\pm$ 0.12}}
& \textbf{32.2 \small{$\pm$ 0.05}} & \textbf{51.9 \small{$\pm$ 0.09}} & \textbf{32.2 \small{$\pm$ 0.05}} \\
\midrule
AdaptiveISP & All
& 56.1 & 72.8 & 60.6
& 32.4 & 52.3 & 32.5 \\
\textbf{Ours} & All
& \textbf{56.4 \small{$\pm$ 0.20}} & \textbf{73.1 \small{$\pm$ 0.15}} & \textbf{60.7 \small{$\pm$ 0.21}}
& \textbf{34.9 \small{$\pm$ 0.45}} & \textbf{55.6 \small{$\pm$ 0.66}} & \textbf{34.8 \small{$\pm$ 0.45}} \\
\bottomrule[1pt]
\end{tabular}
}
\vspace{-0.25cm}
\caption{\textbf{Multi-seed results compared with the strongest baseline (AdaptiveISP~\cite{wang2024adaptiveisp}).}
\MethodName{} consistently outperforms AdaptiveISP across both data conditions and tasks, while maintaining low variance across seeds.}
\vspace{-0.3cm}
\label{tab:multiseed}
\end{table}
\begin{figure}[t]
    \centering
    \includegraphics[width=\linewidth]{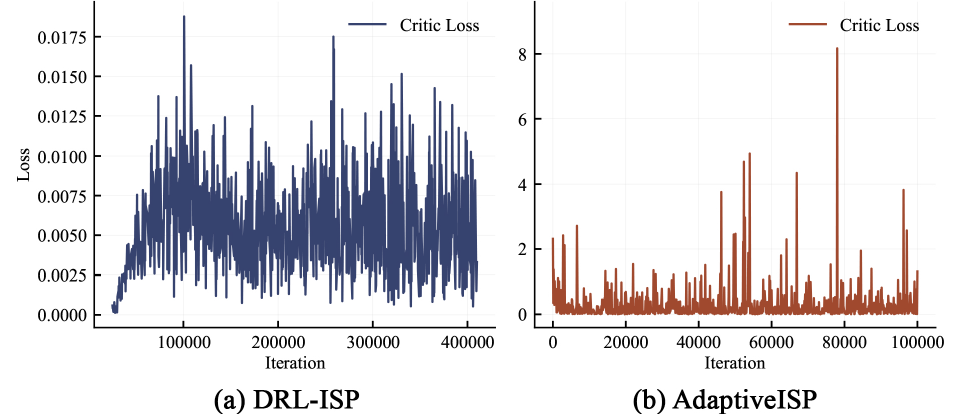}
    \caption{\textbf{Training loss trajectories of critic networks in AdaptiveISP~\cite{wang2024adaptiveisp} and DRL-ISP~\cite{shin2022drlisp}.} The graphs show unstable oscillations and spikes during actor–critic training.}
    \vspace{-0.3cm}
    \label{fig:critic_loss}
\end{figure}
\paragraph{Multi-seed experiments} 
To further analyze training stability, we report results over three random seeds for both object detection and instance segmentation tasks and compare the results with AdaptiveISP~\cite{wang2024adaptiveisp}. 
As shown in~\cref{tab:multiseed}, \MethodName{} consistently outperforms the baseline across different random seeds, indicating stable optimization and reproducible gains.

\paragraph{Instability of the critic network}
\MethodName{} adopts sequence-level optimization to mitigate the unstable optimization behavior observed in prior stepwise RL-based ISP methods~\cite{shin2022drlisp, wang2024adaptiveisp}. 
In these methods, the instability primarily arises from the critic network used to estimate future rewards.
To demonstrate the instability of the critic network adopted by other RL-based methods, we visualize the loss graphs of the critic networks of DRL-ISP~\cite{shin2022drlisp} and AdaptiveISP~\cite{wang2024adaptiveisp} in \cref{fig:critic_loss}. 
As shown in the figure, the learning curves exhibit oscillations and spikes during training, as errors in the bootstrapped critic estimates accumulate and propagate through the policy updates, leading to unstable optimization.

Such oscillatory behavior has also been reported in prior RL works, where bootstrapped value errors and recurrent policy oscillations are identified as key sources of instability~\cite{wagner2011reinterpretation, metelli2021safe, jelley2024efficient}. In contrast, \MethodName{} reformulates the problem as sequence-level optimization, eliminating the need for an unstable critic network.

\subsection{{Runtime Analysis}}

\paragraph{End-to-end runtime}
We analyze the total inference time, including both pipeline estimation and execution, for models trained on the LOD-Dark~\cite{Hong2021Crafting} dataset with a YOLOv3~\cite{redmon2018yolov3} detector. We compare \MethodName{} with DRL-ISP~\cite{shin2022drlisp} and AdaptiveISP~\cite{wang2024adaptiveisp}. The results are reported in \cref{tab:lod_dark_runtime}.
\MethodName{} achieves the best detection performance among the compared methods. 
However, because \MethodName{} predicts ISP pipelines without strict structural constraints, the resulting pipelines tend to be longer, leading to increased inference latency. 
Consequently, its end-to-end runtime is higher than that of prior methods, which explicitly constrain pipeline length.

To reduce this overhead, we can introduce an additional runtime penalty during training, following the strategy used in AdaptiveISP. 
With this penalty, the predicted pipelines become significantly shorter, leading to a substantial reduction in end-to-end latency. 
Importantly, this reduction in runtime is achieved while maintaining detection performance comparable to that of the default \MethodName{} configuration.
These results demonstrate that \MethodName{} provides a flexible mechanism for balancing accuracy and efficiency. By incorporating a runtime-aware training objective, the framework can produce shorter pipelines and improve runtime efficiency without significant degradation in task performance.

\paragraph{On-device runtime analysis}
We further evaluate the runtime on a Galaxy S10 CPU to assess practical deployability.
All measurements are conducted in FP32 precision using a single CPU thread, without quantization or hardware-specific acceleration.
As shown in~\cref{tab:ondevice_runtime}, POS-ISP runs about $4\times$ faster than AdaptiveISP on device. These results demonstrate that \MethodName{} maintains high computational efficiency even on resource-constrained mobile hardware, enabling real-time operation on an older mobile CPU.

\begin{table}[t]
\centering
\resizebox{\columnwidth}{!}{
\begin{tabular}{l|ccc|cc|c}
\toprule
& \multicolumn{3}{c|}{\textbf{LOD-Dark (YOLOv3)}} & \multicolumn{2}{c|}{\textbf{Runtime (ms)}} & \\
\cmidrule(lr){2-4} \cmidrule(lr){5-6}
\textbf{Method} 
& \begin{tabular}{c}mAP\\@0.50{:}0.95\end{tabular}
& \begin{tabular}{c}mAP\\@0.5\end{tabular}
& \begin{tabular}{c}mAP\\@0.75\end{tabular}
& Prediction
& End-to-end
& Length \\
\midrule
DRL-ISP~\cite{shin2022drlisp}     & 44.2 & 67.8 & 48.4 & 15.71 & 26.85 & 3\\
AdaptiveISP~\cite{wang2024adaptiveisp} 
& 47.1 & 71.4 & 51.7 & 12.72 & \underline{$16.30$} & 5 \\

Ours 
& \textbf{47.8} & \textbf{72.1} & \textbf{52.8} & \textbf{1.55} & $82.57$ & 10 \\

Ours (w/ penalty)
& \underline{47.7} & \underline{72.0} & \underline{52.2} & \textbf{1.55} & $\mathbf{1.61}$ & 3 \\

\bottomrule
\end{tabular}}
\vspace{-0.2cm}
\caption{\textbf{Accuracy–runtime trade-off with runtime penalty on LOD-Dark.}
Object detection performance and corresponding runtime measured on a single NVIDIA RTX 2080 Ti with $512\times512$ input images.}
\vspace{-0.1cm}
\label{tab:lod_dark_runtime}
\end{table}

\begin{table}[t]
\centering
\resizebox{\columnwidth}{!}{
\begin{tabular}{lcc}
\toprule
\textbf{Method} & \textbf{Prediction (ms / FPS)} & \textbf{End-to-end (ms / FPS)} \\
\midrule
AdaptiveISP~\cite{wang2024adaptiveisp} & 29.0 / 34.48 & 115.0 / 8.70 \\
Ours (w/ penalty) & \textbf{7.21 / 138.70} & \textbf{32.4 / 30.86} \\
\bottomrule
\end{tabular}
}
\vspace{-0.2cm}
\caption{\textbf{On-device runtime comparison on a Galaxy S10 CPU.}
Runtime is measured using ISP pipelines trained on the LOD-Dark dataset with $512\times512$ input images.}
\vspace{-0.1cm}
\label{tab:ondevice_runtime}
\end{table}

\subsection{{Fixed vs.~Dynamic Pipeline Length}}

{To demonstrate that the performance gains of \MethodName{} stem from its optimization strategy rather than increased pipeline length, we compare it with a fixed-length variant and further examine whether dynamic length adaptation provides additional improvements.}

Following AdaptiveISP~\cite{wang2024adaptiveisp}, which models the pipeline as five processing stages, we train \MethodName{} with the same fixed sequence length and candidate module set.
The results are summarized in \cref{tab:ablation_length}. 
Even under the fixed-length setting, \MethodName{} outperforms AdaptiveISP, indicating that its performance gain arises from sequence-level optimization and stable training guided by final-output rewards, rather than from simply employing a longer pipeline. 
Allowing the pipeline length to be determined dynamically yields further improvements, consistently surpassing both AdaptiveISP and the fixed-length variant.

\begin{table}[t]
\centering
\resizebox{\columnwidth}{!}{
\begin{tabular}{l|c|ccc|ccc}
\toprule
\multirow{2}{*}{\textbf{Method}} & \multirow{2}{*}{\textbf{\shortstack{Length\\Constraint}}} &
\multicolumn{3}{c|}{\textbf{LOD-Dark}} & \multicolumn{3}{c}{\textbf{LIS-Dark}} \\
\cmidrule(lr){3-5}\cmidrule(lr){6-8}
& &
\shortstack{mAP\\@0.5:0.95} & \shortstack{mAP\\@0.5} & \shortstack{mAP\\@0.75} &
\shortstack{mAP\\@0.5:0.95} & \shortstack{mAP\\@0.5} & \shortstack{mAP\\@0.75} \\
\midrule
\multirow{1}{*}{AdaptiveISP~\cite{wang2024adaptiveisp}}
& 5             & 47.1 & {71.4} & \underline{51.7} & 25.2 & 42.3 & 25.2 \\
\midrule
\multirow{2}{*}{Ours} & 5             & \underline{47.4} & \underline{71.7} & {51.6} & \underline{31.0} & \underline{50.7} & \underline{31.1}  \\
& $\le$10 & \textbf{47.8} & \textbf{72.1} & \textbf{52.8} & \textbf{32.1} & \textbf{51.8} & \textbf{32.1}\\

\bottomrule
\end{tabular}
}
\vspace{-0.2cm}
\caption{\textbf{Ablation study on the impact of fixed and dynamic pipeline length.}  We highlight the best metrics as bold and the second best metrics as underline.}
\vspace{-0.3cm}
\label{tab:ablation_length}
\end{table}

\begin{table}[t]
\centering
\small
\resizebox{\columnwidth}{!}{
\begin{tabular}{l|ccc|c|ccc|c}
\toprule
\multirow{3}{*}{\textbf{Method}} 
& \multicolumn{4}{c|}{\textbf{LOD-All}}
& \multicolumn{4}{c}{\textbf{LIS-All}} \\

\cmidrule{2-5} \cmidrule{6-9}

& \multicolumn{3}{c|}{$P(\Delta<0)\downarrow$} 
& \multirow{2}{*}{\shortstack{Worst 5\% \\Mean$\uparrow$}}
& \multicolumn{3}{c|}{$P(\Delta<0)\downarrow$} 
& \multirow{2}{*}{\shortstack{Worst 5\%\\ Mean$\uparrow$}} \\

\cmidrule{2-4} \cmidrule{6-8}

& {All} & {Normal} & {Dark} &
& {All} & {Normal} & {Dark} & \\

\midrule

DRL-ISP~\cite{shin2022drlisp} 
& 0.185 & 0.186 & 0.184 & -0.28
& 0.423 & 0.441 & 0.405 & -0.63 \\

ReconfigISP~\cite{yu2021reconfigisp}
& 0.252 & 0.229 & 0.274 & -0.31
& 0.287 & 0.309 & 0.265 & -0.38 \\

AdaptiveISP~\cite{wang2024adaptiveisp}
& 0.135 & \textbf{0.138} & 0.132 & -0.18
& 0.269 & 0.217 & 0.321 & -0.30 \\

\textbf{Ours}
& \textbf{0.123} & \textbf{0.138} & \textbf{0.108} & \textbf{-0.16}
& \textbf{0.187} & \textbf{0.199} & \textbf{0.174} & \textbf{-0.25} \\

\bottomrule
\end{tabular}
}

\vspace{-0.2cm}
\caption{
\textbf{Output-level robustness analysis using per-image $\Delta$AP@0.5 (w.r.t. input RAW).}
$P(\Delta < 0)$ denotes the fraction of images whose performance degrades after ISP, and Worst 5\% Mean reports the average drop among the most severely degraded cases (AP in the [0,1] scale). Best results are in bold.
}
\label{tab:robustness_analysis}
\vspace{-0.2cm}
\end{table}
\subsection{{Single Sequence per Task}}
\MethodName{} employs a single sequence per task while adapting module parameters on a per-image basis.
In this framework, the sequence defines the overall processing structure of the pipeline, while the parameters control image-specific behavior within that structure.
Although the sequence remains fixed, adapting parameters per image provides sufficient flexibility to accommodate diverse inputs.

Experimental observations support this behavior.
As shown in \cref{tab:robustness_analysis}, our method achieves the lowest degradation rate $P(\Delta < 0)$ and the mildest worst-case drops across both datasets, demonstrating stable behavior across diverse inputs despite using a single sequence.

To further analyze robustness across different illumination conditions, we additionally report $P(\Delta < 0)$ on the Normal (well-lit) and Dark (low-light) subsets in the LOD~\cite{Hong2021Crafting} and LIS~\cite{2023lis} datasets.
Our method shows comparable degradation rates across the two subsets (LOD: $0.138$ vs.\ $0.108$; LIS: $0.199$ vs.\ $0.174$), indicating that the optimized pipeline behaves consistently across lighting variations.

These results indicate that a fixed pipeline sequence combined with per-image parameter adaptation provides sufficient flexibility to handle diverse image conditions while maintaining stable behavior across the dataset.
If a single shared sequence were fundamentally insufficient, we would expect significantly higher degradation rates or substantially worse worst-case drops across subsets, which is not observed in our results.

\begin{table}[t]
\centering
\resizebox{\columnwidth}{!}{
\begin{tabular}{l|ccc|ccc}
\toprule[1pt]
& \multicolumn{3}{c|}{\textbf{LOD-Dark}} 
& \multicolumn{3}{c}{\textbf{LOD-All}} \\
\cmidrule(lr){2-4}\cmidrule(lr){5-7}
\multirow{-3}{*}{\textbf{Method}} 
& \shortstack{mAP\\@0.5:0.95} & \shortstack{mAP\\@0.5} & \shortstack{mAP\\@0.75} 
& \shortstack{mAP\\@0.5:0.95} & \shortstack{mAP\\@0.5} & \shortstack{mAP\\@0.75}  \\
\midrule
Input RAW   & 34.8 & 53.8 & 37.5 & 44.3 & 62.2 & 46.5  \\
Camera ISP  & 26.9 & 41.4 & 28.9 & 40.1 & 56.2 & 42.4 \\
DRL-ISP~\cite{shin2022drlisp} & 35.4 & 54.7 & 38.0 & 44.7 & 62.5 & 46.7 \\
ReconfigISP~\cite{yu2021reconfigisp} & 33.4 & 52.0 & 35.9 & 14.0 & 20.7 & 14.6 \\
AdaptiveISP~\cite{wang2024adaptiveisp} & 36.8 & 56.1 & 40.2 & 47.3 & 65.4 & 49.9 \\
\textbf{Ours} & \textbf{38.1} & \textbf{57.9} & \textbf{41.7} & \textbf{48.1} & \textbf{65.5} & \textbf{51.0} \\
\bottomrule[1pt]
\end{tabular}
}
\vspace{-0.2cm}
\caption{\textbf{Quantitative comparison with object detection task with YOLOv13.} In the table, ``Input RAW'' denotes images that have undergone only the most basic preprocessing steps: defective pixel correction, black level correction, lens shading correction, and demosaicking. We highlight the best metrics as bold.}
\vspace{-0.1cm}
\label{tab:comp_detect_v13}
\end{table}

\subsection{{Generalization Across Optimization Settings}}
To evaluate the robustness of \MethodName{} beyond a specific optimization setting, we analyze its behavior across different reward backbones and training objectives.

\paragraph{Reward backbone}
We evaluate generalization across different detection backbones used for reward computation. While the main paper computes task rewards using a single pretrained detector, we conduct additional experiments with YOLOv13~\cite{lei2025yolov13}. As reported in \cref{tab:comp_detect_v13}, ReconfigISP~\cite{yu2021reconfigisp} performs worse than the input images due to the mismatch between soft training and hard inference. On LOD-All, the optimized pipeline tends to overfit the Dark subset, resulting in overly aggressive enhancement that causes Normal images to become saturated. Because this single fixed pipeline is applied uniformly to all inputs without adapting to individual image characteristics, the performance degradation becomes significant. Other RL-based methods~\cite{shin2022drlisp, wang2024adaptiveisp} improve upon the input images but remain suboptimal, as their stepwise optimization depends on unstable estimation of future rewards. In contrast, \MethodName{} achieves the best performance with YOLOv13, demonstrating strong generalization across different pretrained reward models.
Overall, these results show that sequence-level joint optimization provides a stable and backbone-agnostic framework that remains effective under varying training objectives and reward models.
\vspace{-0.3cm}

\paragraph{Training objective}
To evaluate robustness to different training objectives, we train \MethodName{} and AdaptiveISP~\cite{wang2024adaptiveisp} on the depth estimation task using different loss functions, including Root Mean Squared Error (RMSE), Absolute Difference (AbsDiff), and Absolute Relative Error (AbsRel), under the same SC-SfMLearner~\cite{bian2019unsupervised} supervision. These objectives introduce different optimization biases: RMSE emphasizes large depth errors, AbsDiff measures average absolute deviation, and AbsRel captures scale-aware discrepancies.

As shown in \cref{tab:loss_generalization}, \MethodName{} consistently outperforms AdaptiveISP across all training objectives. The performance gains remain stable across different loss functions and evaluation metrics, indicating that our method does not depend on a specific objective and exhibits robust optimization behavior under diverse depth supervision signals.
\begin{table}[t]
\centering
\resizebox{0.95\columnwidth}{!}{
\begin{tabular}{c|l|cccc}
\toprule
\multirow{2}{*}{\textbf{Loss}} 
& \multirow{2}{*}{\textbf{Method}} 
& \multicolumn{3}{c}{\textbf{Error} $\downarrow$} 
& \textbf{Accuracy} $\uparrow$ \\
\cmidrule(lr){3-5} \cmidrule(lr){6-6}
&  & AbsDiff & AbsRel & RMS & $\delta < 1.25$ \\
\midrule

\multirow{2}{*}{RMSE}
& AdaptiveISP 
& 2.574 & 0.131 & 5.22 & 0.840 \\
& \textbf{Ours}
& \textbf{2.525} & \textbf{0.128} & \textbf{5.13} & \textbf{0.847} \\

\midrule

\multirow{2}{*}{AbsDiff}
& AdaptiveISP
& 2.548 & 0.128 & 6.24 & 0.847 \\
& \textbf{Ours}
& \textbf{2.504} & \textbf{0.127} & \textbf{5.15} & \textbf{0.850} \\

\midrule

\multirow{2}{*}{AbsRel}
& AdaptiveISP
& 2.656 & 0.134 & 5.38 & 0.833 \\
& \textbf{Ours}
& \textbf{2.522} & \textbf{0.126} & \textbf{5.19} & \textbf{0.847} \\

\bottomrule
\end{tabular}
}
\vspace{-0.1cm}
\caption{\textbf{Comparison under different training objectives.}
For each objective setting, \MethodName{} consistently outperforms AdaptiveISP~\cite{wang2024adaptiveisp} across all metrics.}
\vspace{-0.2cm}
\label{tab:loss_generalization}
\end{table}

\subsection{{Ablation on Parameter Predictor}}
We observe that conditioning the parameter predictor solely on the input image is sufficient in our training setting. Early in training, the sequence policy exhibits high entropy and explores a wide range of ISP pipelines. In this regime, an image-only predictor (IO) learns parameters over transient and rapidly changing sequences. As training progresses, the policy entropy decreases and the sequence distribution gradually concentrates on a small set of high-reward pipelines. Consequently, the predictor ultimately needs to model only the final converged pipeline, making image-only conditioning adequate in practice.

For comparison, we implement a sequence-conditioned variant (SC), where sequence features are modulated by image features and the latent dimension is increased by a factor of two, while keeping the rest of the architecture unchanged. Although SC is more expressive in principle, rapidly changing sequences during early training introduce unstable conditioning that interferes with parameter learning. As shown in \cref{tab:param_predictor_ablation}, SC exhibits higher variance and lower performance across random seeds compared to IO. These results suggest that additional conditioning complexity in the predictor does not necessarily improve optimization in our sequence-level training framework.

\begin{table}[t]
\centering
\resizebox{\columnwidth}{!}{
\begin{tabular}{l|ccc|ccc}
\toprule
\multicolumn{1}{c|}{\multirow{2}{*}{\shortstack{Parameter\\Predictor}}} &
\multicolumn{3}{c|}{\textbf{LOD-Dark}} &
\multicolumn{3}{c}{\textbf{LIS-Dark}} \\
\cmidrule(lr){2-4} \cmidrule(lr){5-7}
& \shortstack{mAP\\@0.5:0.95} & \shortstack{mAP\\@0.5} & \shortstack{mAP\\@0.75}
& \shortstack{mAP\\@0.5:0.95} & \shortstack{mAP\\@0.5} & \shortstack{mAP\\@0.75} \\
\midrule
(a) SC
& 47.5 $\pm$ 0.26 & 71.5 $\pm$ 0.28 & 52.0 $\pm$ 0.42
& 31.6 $\pm$ 0.33 & 51.2 $\pm$ 0.42 & 31.3 $\pm$ 0.52 \\
(b) IO (Ours)
& \textbf{47.8 $\pm$ 0.08} & \textbf{72.1 $\pm$ 0.16} & \textbf{52.7 $\pm$ 0.12}
& \textbf{32.2 $\pm$ 0.05} & \textbf{51.9 $\pm$ 0.09} & \textbf{32.2 $\pm$ 0.05} \\
\bottomrule
\end{tabular}}
\vspace{-0.2cm}
\caption{
\textbf{Comparison between sequence-conditioned (SC) and image-only (IO) parameter predictors.} 
We report mean $\pm$ standard deviation over three seeds.
}
\vspace{-0.3cm}
\label{tab:param_predictor_ablation}
\end{table}

\subsection{Importance of Task-adaptive Sequence}
\begin{table}[t]
\centering
\resizebox{\columnwidth}{!}{
\begin{tabular}{l|ccc}
\toprule[1pt]
\textbf{Source Task (Sequence)} &
mAP@0.5:0.95 &
mAP@0.5 &
mAP@0.75 \\
\midrule
Image Enhancement   & 47.6 & 71.7 & 52.4 \\
Instance Segmentation  & 47.1 & 71.4 & 50.9 \\
Object Detection & \textbf{48.0} & \textbf{72.2} & \textbf{52.8} \\
\bottomrule[1pt]
\end{tabular}
}
\vspace{-0.2cm}
\caption{
\textbf{Cross-task evaluation.} Module sequences from different source tasks, evaluated on object detection with parameters retrained for that task. We highlight the best metrics as bold.
}
\vspace{-0.3cm}
\label{tab:cross_task_pipeline}
\end{table}

We train the \ActionNetName{} to adapt its sequence to the target task.
To assess the importance of such task-aware sequences in downstream task performance, we conduct an experiment.
In this experiment, we first obtain three sequences that are optimized for image enhancement, instance segmentation, and object detection, respectively.
Then, we construct three ISP pipelines using the sequences and optimize only the \ParamNetName{} on the object detection task using the LOD-Dark~\cite{Hong2021Crafting} dataset and YOLOv3~\cite{redmon2018yolov3} detector.
By optimizing only the module parameters while keeping the module sequences fixed, we can examine the effect of the module sequence on the overall ISP pipeline.

The results are reported in \cref{tab:cross_task_pipeline}.
As shown in the table, the sequence optimized for other tasks yields noticeably lower performance when applied to the detection task, even after its parameters are retrained. 
In contrast, the sequence originally optimized for the same task, object detection, preserves strong performance. 
These results indicate that the module sequence itself plays a decisive role in downstream performance and cannot be recovered through parameter optimization alone. 
This finding underscores the necessity of task-aware sequence design, as different tasks naturally favor different ISP module orders.

\subsection{Image-adaptive Parameter Prediction}
\begin{table}[t]
\centering
\resizebox{0.95\columnwidth}{!}{
\begin{tabular}{ll|ccc}
\toprule[1pt]
\textbf{Estimated from} & \textbf{Applied to} & 
\shortstack{mAP@0.5:0.95} & \shortstack{mAP@0.5} & 
\shortstack{mAP@0.75} \\
\midrule
LOD-Normal   & LOD-Dark  & 52.6  & 70.0 & 57.0 \\
LOD-Dark   & LOD-Dark  & \textbf{53.6}  & \textbf{71.2} & \textbf{57.4} \\
\bottomrule
\end{tabular}
}
\vspace{-0.2cm}
\caption{
\textbf{Cross-exposure evaluation of ISP parameter transfer between different datasets.} We highlight the best metrics as bold.}
\vspace{-0.2cm}
\label{tab:ablation_params}
\end{table}

\begin{figure}[t]
    \centering
    \includegraphics[width=1.0\linewidth]{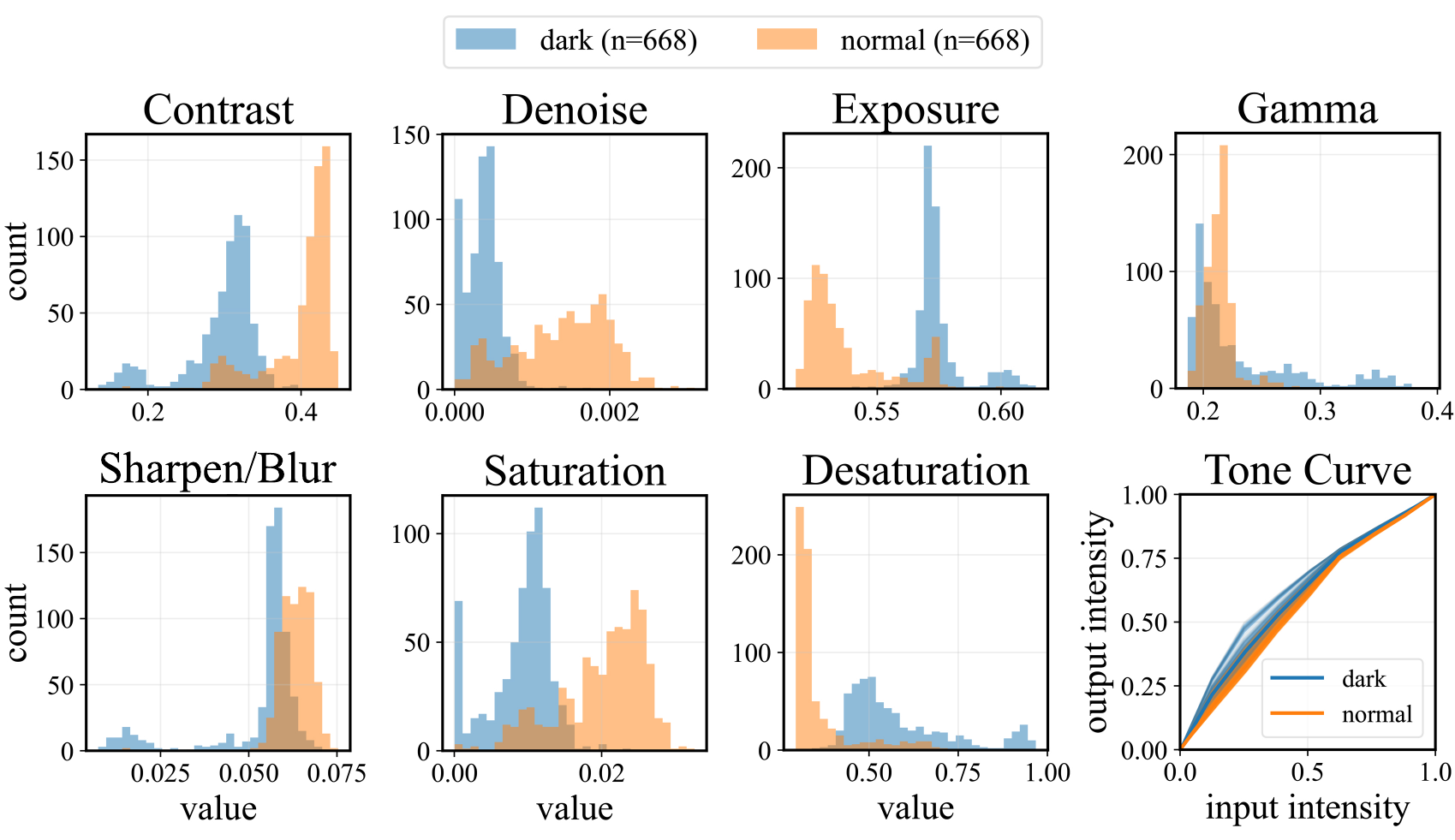}
    \caption{\textbf{Histogram comparisons of the predicted parameters across illumination domains.}}
    \label{fig:hist_param}
\end{figure}
\begin{figure}[t]
    \centering
    \includegraphics[width=\linewidth]{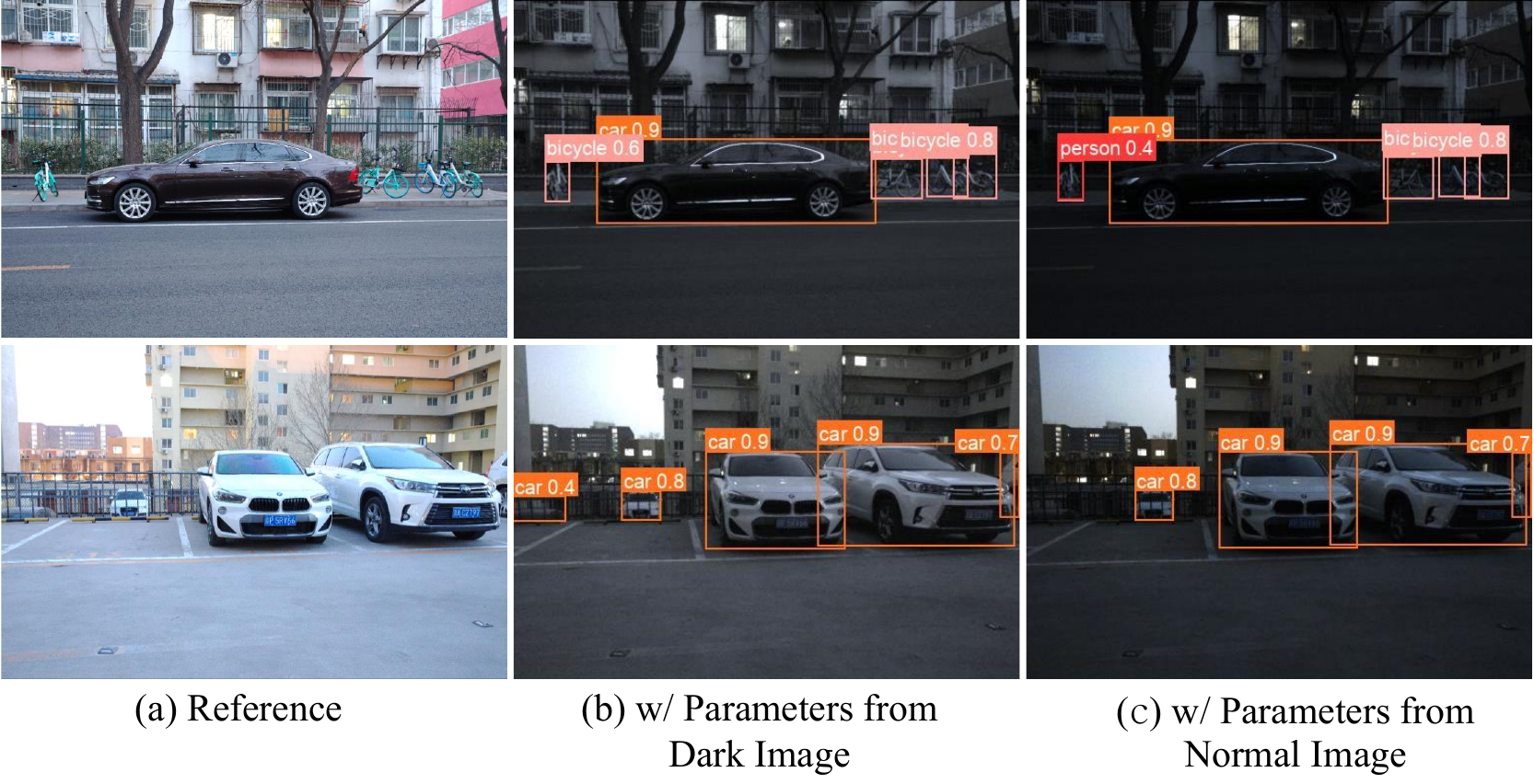}
    \vspace{-0.5cm}
    \caption{\textbf{Qualitative examples of cross-domain parameter swapping.} Reference images are well-lit scenes from the LOD-Normal dataset, with brightness increased by $1.5\times$ for visualization. ISP parameters are estimated from different domains and applied to LOD-Dark images.} 
    \vspace{-0.2cm}
    \label{fig:param_qual}
\end{figure}

Our \ParamNetName{} conditions on the input image to produce image-adaptive module parameters.
To examine whether these parameters truly respond to scene characteristics, particularly illumination, we design a cross-domain parameter-swap experiment.
We use our \MethodName{} model trained on LOD-All~\cite{Hong2021Crafting} with the YOLOv3-based~\cite{redmon2018yolov3} detection loss.
Using this model, we estimate ISP parameters from images in either LOD-Dark or LOD-Normal.
We then apply these parameters to LOD-Dark images and evaluate the resulting  performance.

\cref{tab:ablation_params} shows the quantitative results.
When parameters are predicted and applied within the same image domain, the model achieves the highest performance in each case.
However, when parameters are transferred across domains, performance consistently drops.
This degradation confirms that the predicted parameters are sensitive to scene conditions and that the \ParamNetName{} effectively adapts its predictions to the input image.

We additionally visualize the distributions of the predicted parameters in~\cref{fig:hist_param} as well as the output images in~\cref{fig:param_qual}. 
When parameters estimated from LOD-Normal images are applied to LOD-Dark scenes, the resulting outputs become underexposed or locally flattened, clearly reflecting the domain mismatch.
The parameter histograms also reveal noticeable shifts between Dark and Normal inputs: exposure- and tone-related parameters predicted for LOD-Dark images tend to apply stronger brightness compensation, whereas those predicted for LOD-Normal images show more moderate adjustments. 
This separation in parameter distributions indicates that the parameter predictor does not regress to a domain-agnostic average but instead adapts its predictions to the illumination characteristics of each input. 
Together with the performance drop observed under cross-domain transfer, these qualitative and statistical findings further demonstrate that \MethodName{} learns image-aware parameter prediction closely aligned with scene conditions.

\section{Implementation Details}
\subsection{ISP Modules}
\begin{table*}[t]
\centering
\scalebox{0.9}{
\begin{tabular}{@{} l l l c l @{}}
\toprule
\textbf{Module} & \textbf{Equation} & \textbf{Parameter Type} & \textbf{\# Params} & \textbf{Parameter Range} \\
\midrule

Exposure control &
$\mathbf{I}' = 2^{\theta_e'}\, \mathbf{I}$ &
EV shift $\theta_e'$ & 1 &
$\theta_e' \in [-3.5, 3.5]$ \\

Gamma correction &
$\mathbf{I}' = \mathbf{I}^{\,\theta_g'}$ &
Gamma exponent $\theta_g'$ & 1 &
$\theta_g' \in [\tfrac{1}{3}, 3]$ \\

Tone mapping &
$\mathbf{I}' = \frac{8}{Z} \sum_{i=1}^{8} \theta_{t,i}'\, b_i(\mathbf{I})$ &
Segment weights $\theta_{t,i}'$ & 8 &
$\theta_{t,i}' \in [0.5, 2.0]$ \\

Contrast &
$\mathbf{I}' = (1-\theta_c')\, \mathbf{I} + \theta_c'\, \tilde{\mathbf{I}}$ &
Contrast factor $\theta_c'$ & 1 &
$\theta_c' \in [-1, 1]$ \\

Saturation &
$\mathbf{I}' = (1-\theta_s)\, \mathbf{I} + \theta_s\, \mathbf{I}_{\mathrm{enh}}$ &
Blend factor $\theta_s$ & 1 &
$\theta_s \in [0, 1]$ \\

Desaturation &
$\mathbf{I}' = (1-\theta_d)\, \mathbf{I} + \theta_d\, \mathbf{I}_{\mathrm{gray}}$ &
Blend factor $\theta_d$ & 1 &
$\theta_d \in [0, 1]$ \\

White balance &
$\mathbf{I}' =
\mathrm{diag}(\tilde{\theta}_{wb,R}',\, \tilde{\theta}_{wb,G}',\, \tilde{\theta}_{wb,B}')\, \mathbf{I}$ &
Normalized gains $\tilde{\theta}_{wb,c}'$ & 3 &
$\theta_{wb,c}' \in [1/1.1,\, 1.1]$ \\

Denoise &
$\mathbf{I}' =
\frac{\sum_{y \in \mathcal{N}(x)}
w(x,y \mid \theta_d')\, \mathbf{I}(y)}
{\sum_{y \in \mathcal{N}(x)} w(x,y \mid \theta_d')}$ &
Denoise strength $\theta_d'$ & 1 &
$\theta_d' \in [10^{-5}, 1]$ \\

Sharpen/Blur &
$\mathbf{I}' = \theta_{sh}'\, \mathbf{I} + (1-\theta_{sh}')\, \mathbf{I}_{\text{blur}}$ &
Sharpen/blur factor $\theta_{sh}'$ & 1 &
$\theta_{sh}' \in [10^{-5}, 10]$ \\

Color Correction &
$\mathbf{I}' = M\, \mathbf{I}$ &
Color correction matrix entries $\theta_{ccm}'$ & 9 &
$\theta_{ccm,ij}' \in [-2, 2]$ \\

\bottomrule
\end{tabular}}
\caption{\textbf{Details of the adopted ISP modules.}}
\label{tab:isp_module_summary}
\end{table*}

The \ParamNetName{} predicts parameter vectors for the ISP modules, where the dimensionalities are module-specific and the elements are normalized to the range of $[0, 1]$.
For a module $\mathcal{M}_i$, we denote its parameters by $\theta_i \in [0, 1]^{d_i}$, where $d_i$ is the number of parameters for that module.
These normalized parameters are then rescaled to their corresponding module-specific numerical ranges and used to parameterize the modules.

Following AdaptiveISP~\cite{wang2024adaptiveisp}, we have 10 candidate modules from which modules are selected to construct an ISP pipeline.
In the following, we describe the modules that constitute our module pool, their formulations, and the parameter ranges of each module.

In the following section, we denote an RGB pixel $\mathbf{I}$ as
\begin{equation}
    \mathbf{I} = (I_R,\, I_G,\, I_B)^\top,
\end{equation}
where $I_R$, $I_G$, and $I_B$ represent the intensities of the red, green, and blue channels, respectively.
We also denote an output RGB pixel of a module as $\mathbf{I'}$.

\paragraph{Exposure control}
For a module for exposure control, the \ParamNetName{} predicts $\theta_e \in [0, 1]$.
The parameter $\theta_e$ is then rescaled to $\theta_e' \in [-3.5, 3.5]$ and used to adjust the brightness of the input pixel $\mathbf{I}$:
\begin{equation}
    \mathbf{I}' = 2^{\theta_e'}\, \mathbf{I}.
\end{equation}

\paragraph{Gamma correction}
The \ParamNetName{} predicts $\theta_g \in [0, 1]$ for the gamma correction module.
The parameter $\theta_g$ is then rescaled to $\theta_g' \in \left[\tfrac{1}{3}, 3\right]$, which controls the gamma curve for nonlinear brightness shaping:
\begin{equation}
    \mathbf{I}' = {\mathbf{I}^{\,\theta_g'}}.
\end{equation}

\paragraph{Tone mapping}
We first define eight piecewise-linear basis functions that construct a tone-mapping curve:
\begin{equation}
    b_i(u)
    = \max\!\left( 0,\; \min\!\left(u - \frac{i-1}{8},\, \frac{1}{8}\right) \right),
\end{equation}
where $i = 1, ..., 8$.
For a module for tone mapping, the \ParamNetName{} predicts $\theta_t \in [0, 1]^8$, which is rescaled to $\theta_t' \in [0.5, 2.0]^8$.
Then, the tone mapping curve is applied as:
\begin{equation}
    \mathbf{I}' = \frac{8}{Z} \sum_{i=1}^{8} \theta_{t,i}'\, b_i(\mathbf{I}),
    \qquad Z = \sum_{i=1}^{8} \theta_{t,i}',
\end{equation}
where $\theta_{t,i}'$ is $i$-th element of $\theta_t'$.

\paragraph{Contrast}
The \ParamNetName{} predicts a parameter $\theta_c \in [0,1]$, which is
rescaled to $\theta_c' \in [-1,1]$ and used as a contrast-adjustment
factor. The module enhances contrast by blending the original RGB vector
with an S-curve–reshaped version of the image.

The luminance is computed as
\begin{equation}
    I_{\mathrm{lum}}
    =
    0.27 I_R + 0.67 I_G + 0.06 I_B,
\end{equation}
and its S-curve transformation and RGB scaling are defined as
\begin{equation}
    \tilde{\mathbf{I}}
    =
    \mathbf{I}
    \cdot
    \frac{I_{\mathrm{S}}}{I_{\mathrm{lum}} + \varepsilon},
    \qquad
    I_{\mathrm{S}}
    =
    \frac{1 - \cos(\pi I_{\mathrm{lum}})}{2},
\end{equation}
where $\varepsilon$ prevents division by zero.
Finally, the contrast-adjusted output is
\begin{equation}
    \mathbf{I}_{\text{contrast}}
    =
    (1 - \theta_c')\, \mathbf{I}
    +
    \theta_c'\, \tilde{\mathbf{I}}.
\end{equation}

\paragraph{Saturation}
To control saturation, we first convert $\mathbf{I}$ into HSV color space, yielding
\begin{equation}
    \mathbf{I}_\text{HSV} = (H, S, V)^\top.
\end{equation}
Then, we strengthen the saturation via
\begin{equation}
    S_{\mathrm{enh}}
    = S + 0.8\cdot(1-S)(0.5 - |0.5 - V|)\, .
\end{equation}
The saturated pixel is converted to RGB color space again, yielding
\begin{equation}
    \mathbf{I}_{\text{enh}} = (I_R^{enh} ,I_G^{enh}, I_B^{enh})^\top.
\end{equation}
The output pixel is then blended with the saturated pixel as following:
\begin{equation}
    \mathbf{I}' = (1-\theta_s)\, \mathbf{I} + \theta_s\, \mathbf{I}_{\mathrm{enh}},
\end{equation}
where $\theta_s \in [0, 1]$ is predicted by the \ParamNetName{}.

\paragraph{Desaturation}
This module reduces colorfulness by blending the RGB vector with its grayscale version. The \ParamNetName{} predicts a parameter $\theta_d \in [0,1]$, which controls how strongly the pixel is pulled toward a monochrome appearance.

The luminance and the corresponding grayscale RGB vector are defined as
\begin{equation}
\begin{aligned}
    I_{\text{lum}} &= 0.27 I_R + 0.67 I_G + 0.06 I_B, \\
    \mathbf{I}_{\text{gray}}
    &= (I_{\text{lum}},\, I_{\text{lum}},\, I_{\text{lum}})^\top,
\end{aligned}
\end{equation}
and the output is produced by linearly interpolating between the original
color and its grayscale representation:
\begin{equation}
    \mathbf{I}'
    = (1 - \theta_d)\, \mathbf{I}
      + \theta_d\, \mathbf{I}_{\text{gray}}.
\end{equation}

\paragraph{White balance}
For white balancing, the \ParamNetName{} predicts $\theta_{wb} \in [0, 1]^3$ which is rescaled to $\theta_{wb}' \in [\displaystyle\frac{1}{1.1}, 1.1]^3$. These rescaled parameters are then converted into the final per-pixel gain values through a luminance-preserving normalization. 
With $\theta_{wb}' = (\theta_{wb,R}',\, \theta_{wb,G}',\, \theta_{wb,B}')^\top$, we compute gain values using:
\begin{equation}
    \tilde{{\theta}}_{wb}'
    =
    \frac{{\theta}_{wb}'}{
        0.27\, \theta_{wb,R}'
        + 0.67\, \theta_{wb,G}'
        + 0.06\, \theta_{wb,B}'
    }.
\end{equation}
The white balanced pixel is then obtained by applying the diagonal matrix of normalized gains:
\begin{equation}
    \mathbf{I}'
    =
    \begin{bmatrix}
        \tilde{\theta}_{wb,R}' & 0 & 0 \\
        0 & \tilde{\theta}_{wb,G}' & 0 \\
        0 & 0 & \tilde{\theta}_{wb,B}'
    \end{bmatrix}
    \mathbf{I}.
\end{equation}

\paragraph{Denoise}
The \ParamNetName{} predicts a parameter $\theta_{d} \in [0,1]$, which is
mapped to a denoising strength $\theta_{d}' \in [10^{-5}, 1]$.
This value controls the smoothing level of the non-local means (NLM) operator.

For each pixel, the denoised value is computed as a weighted average of
neighboring pixels:
\begin{equation}
    \mathbf{I}'
    =
    \frac{\sum_{y \in \mathcal{N}(x)}
        w(x,y \mid \theta_{d}')\, \mathbf{I}(y)}
        {\sum_{y \in \mathcal{N}(x)} w(x,y \mid \theta_{d}')},
\end{equation}
where $y$ indexes spatial locations in the neighborhood $\mathcal{N}(x)$,
and $w(x,y \mid \theta_{d}')$ denotes the NLM weight between pixel $x$
and pixel $y$, scaled according to the denoising strength $\theta_{d}'$.

\paragraph{Sharpen/Blur}
For a module that applies sharpening or blurring operation, the \ParamNetName{} predicts $\theta_{sh} \in [0, 1]$ which is rescaled to $\theta_{sh}' \in [10^{-5}, 10]$. We then define a $3\times3$ blur kernel $K$
\begin{equation}
    K = \frac{1}{13}
    \begin{bmatrix}
        1 & 1 & 1 \\
        1 & 5 & 1 \\
        1 & 1 & 1
    \end{bmatrix},
\end{equation}
to produce a blurry pixel $\mathbf{I}_\text{blur}$ by convolving $K$ with the input image.
The final output is then obtained by
\begin{equation}
    \mathbf{I}' = \theta_{sh}'\, \mathbf{I} + (1-\theta_{sh}')\,\mathbf{I}_{\text{blur}}.
\end{equation}

\paragraph{Color correction}
For color correction, the \ParamNetName{} predicts $\theta_{ccm} \in [0, 1]^9$ which is rescaled to $\theta_{ccm}' \in [-2, 2]^9$.
Then, we define a color correction matrix $M$ with the nine elements of $\theta_{ccm}'$ and normalize each row.
With the color correction matrix, we apply:
\begin{equation}
    \mathbf{I}' = M\, \mathbf{I}.
\end{equation}

\subsection{Network Architecture Details}
\subsubsection{Sequence Predictor}
We instantiate the sequence-level policy with a GRU-based autoregressive model that predicts ISP module indices.
Let $A = (a_1, \dots, a_T)$ denote a module sequence, where $T \le M + 1$. 
Here, $M$ is the number of available ISP modules, and we additionally include a special end-of-sequence token \texttt{<eos>}. 
Thus, the module pool contains $M$ actual modules plus one \texttt{<eos>} token, and the sequence length $T$ can be up to $M+1$ when \texttt{<eos>} appears in the final position.
At each recurrent step $t$, the previously predicted module index $a_{t-1}$ is embedded by a learnable token embedding layer, and passed through a GRU. Then, the GRU produces a hidden state $h_t \in \mathbb{R}^{H}$ from the embedded token where $H=128$ in our implementation.

To provide the decoder with an explicit notion of the step index for step-dependent behavior, we additionally use a learnable step embedding $s_t$ for $t \in {1,\dots,T}$ of dimension 16, following FiLM~\cite{perez2017filmvisualreasoninggeneral}.
This step embedding is transformed into FiLM parameters $(\gamma_t, \beta_t)$ through a small MLP and is used to modulate the hidden state $h_t$ as
\begin{equation}
\tilde{h}_t = h_t \odot (1 + \gamma_t) + \beta_t,
\end{equation}
where $\odot$ denotes element-wise multiplication.
The modulated hidden state $\tilde{h}_t$ is then projected by a linear decoder to produce the logits over module indices and the \texttt{<eos>} token.
To avoid introducing any bias toward specific modules at the beginning of training, we initialize the decoder such that all logits are identical: the weight matrix is set to zero and all bias entries are set to the same constant, so that the initial policy corresponds to a uniform probability distribution over actions after the softmax.

We sample the next action from the resulting probability distribution with a temperature-controlled softmax; during training, the temperature $\tau$ follows an exponentially decaying schedule to gradually reduce exploration.
To avoid repeated selection of modules, we mask out the already selected modules in the logit space so that each module can be chosen at most once.
The decoding process terminates when the \texttt{<eos>} token is selected.

\paragraph{Temperature sampling}
To balance exploration and exploitation during training, we decay the sampling temperature~$\tau$ using an exponential schedule. 
The temperature scales the logits as $\mathrm{softmax}(z/\tau)$, where a higher~$\tau$ yields a smoother distribution for broader exploration.
As training progresses, $\tau$ is gradually reduced, producing a sharper distribution that favors high-reward actions.
At inference time, exploration is disabled, and the module with the highest probability is selected.

We set $\tau_{\max}=2.5$ and $\tau_{\min}=0.2$, and update $\tau$ at training step~$t$ as
\[
\tau(t)
= \tau_{\min} 
  + \bigl(\tau_{\max} - \tau_{\min}\bigr)
    \exp\!\left(-\ln 2 \cdot \frac{t}{h}\right),
\]
where $h = 3000$ denotes the half-life in training steps. This schedule halves the gap $\tau(t) - \tau_{\min}$ every $h$ steps, providing a smooth shift from exploration to more deterministic behavior as training stabilizes.

\subsubsection{Parameter Predictor}
The parameter predictor is implemented as a lightweight encoder-decoder network that maps the input image to a compact latent vector and then to a concatenated parameter vector for all modules.
Given an input image $I \in \mathbb{R}^{3 \times H \times W}$, we first downsample it to a fixed spatial resolution of $64 \times 64$ using adaptive average pooling.
The downsampled image is then passed through three convolutional blocks with LeakyReLU activations, resulting in a feature map $F \in \mathbb{R}^{4C \times 8 \times 8}$.
We apply both global average pooling and global max pooling to $F$ to obtain two separate pooled vectors, and then concatenate them to form an image feature vector of dimension $8C$.

This vector is passed through a two-layer MLP, where the first fully connected layer reduces the dimension to 256 with a ReLU activation, and the second layer projects it to a latent vector $z \in \mathbb{R}^{D}$, where $D$ is the latent dimension.
Finally, a small decoder MLP maps $z$ to the output parameter vector of dimension $P$ that contains continuous parameters for all modules. The parameters corresponding to the selected modules are then retrieved from this vector.
We then apply a \texttt{tanh} activation and rescale the output from the range $[-1, 1]$ to $[0, 1]$.

\subsection{{Details on Running Other Methods}}
\paragraph{DRL-ISP}
DRL-ISP~\cite{shin2022drlisp} is originally implemented to process images with four channels of $(R, G_1, G_2, B)$.
For datasets in RGB format with three channels, it duplicates the G channel to construct four channels of $(R, G, G, B)$.
Since the datasets we use provide only RGB images, we duplicate the green channel to form a four-channel representation $(R, G, G, B)$ following DRL-ISP. Using these four-channel images, we retrain the proxy networks of DRL-ISP so that they operate correctly on this input format. During ISP pipeline optimization with the proxy networks, we also use the same four-channel $(R, G, G, B)$ images.

\paragraph{ReconfigISP} 
ReconfigISP~\cite{yu2021reconfigisp} includes several RAW-domain ISP modules, such as RAW\,$\rightarrow$\,RAW denoising (Bilateral-Bayer, Median-Bayer, NLM-Bayer, Path-Restore-Bayer) and RAW\,$\rightarrow$\,sRGB demosaicking (Laplacian, Nearest, Bilinear, DemosaicNet), all of which are designed for Bayer inputs. 
We exclude the modules that operate in the RAW domain since our data are entirely in the RGB domain.
We then use only the modules that operate in the RGB domain.
Moreover, since the original ReconfigISP fixes the number of RGB modules to three, we also set our pipeline length to three for consistency.

\paragraph{AdaptiveISP} 
AdaptiveISP~\cite{wang2024adaptiveisp} operates under the same input assumptions as our method; therefore, we employ their default configuration without any additional modifications.

\subsection{{Details on Dataset}}
\paragraph{Adobe FiveK Dataset}
In the Adobe FiveK dataset~\cite{fivek}, the RAW images and the expert-retouched images are not geometrically aligned because the experts process the images with Adobe Lightroom, whose internal pipeline introduces subtle geometric transformations (e.g., resampling, slight cropping, or perspective adjustments).
Because our training relies on pixel-wise loss computation, the input and target images need to be geometrically aligned.
To ensure this alignment, we process the original RAW images through the Adobe Lightroom pipeline and then convert the results to linear RGB. 
The overall conversion procedure follows exactly the method in \cite{hu2018exposure}. {We train the models on 3,000 images and evaluate them on 1,000 test images.}

\paragraph{LOD Dataset}
{The LOD dataset~\cite{Hong2021Crafting} provides two subsets: LOD-Normal (well-lit) and LOD-Dark (low-light).
For LOD-Dark, we follow the same experimental protocol as prior task-driven ISP works, including AdaptiveISP~\cite{wang2024adaptiveisp}, and use the identical data split. LOD-Dark consists of 1,781 training images and 400 test images.}
{
In addition, we evaluate on LOD-All, which combines both Normal and Dark subsets. LOD-All contains 3,122 training images and 1,336 test images, where both the training and test splits include an equal number of Normal and Dark images.}

\paragraph{LIS Dataset}
{The LIS dataset~\cite{2023lis} provides segmentation annotations for instance-level evaluation. In our experiments, LIS-Dark contains 1,561 training images and 669 test images, and LIS-All includes 3,122 training images and 1,338 test images, each split having an equal number of Normal and Dark images.}

\paragraph{KITTI Dataset} {For monocular depth estimation, we use the KITTI dataset~\cite{geiger2012we}. Following DRL-ISP~\cite{shin2022drlisp}, we adopt their synthetic RAW version of KITTI, which is generated by converting the original RGB images into RAW Bayer images. The resulting dataset consists of 37,430 training images and 697 test images. We apply demosaicing to the synthetic RAW images and use the resulting images as inputs to our ISP optimization framework.}

{
    \small
    \bibliographystyle{ieeenat_fullname}
    \bibliography{main}
}
